\documentclass[runningheads]{llncs}
\usepackage{graphicx}
\usepackage[dvipsnames]{xcolor}

\usepackage{tikz}
\usepackage{comment}
\usepackage{amsmath,amssymb} 
\usepackage{color}
\usepackage{multirow}

\usepackage[normalem]{ulem}
\usepackage[dvipsnames]{xcolor}

\usepackage{float}
\usepackage{nicefrac}
\usepackage{footnote}
\usepackage{xspace}
\usepackage{booktabs}
\usepackage{cite}
\usepackage{array}

\usepackage{subcaption}
\usepackage[labelfont=bf, skip=1pt]{caption}

\usepackage[pagebackref,breaklinks,colorlinks]{hyperref}

\usepackage{pifont}
\newcommand{\cmark}{\textcolor{ForestGreen}{\ding{51}}}%
\newcommand{\xmark}{\textcolor{Red}{\ding{55}}}%

\makesavenoteenv{tabular}
\makesavenoteenv{table}

\newcommand\our[1][]{CRCT\xspace}
\newcommand\ourFullname[1][]{Classification - Regression Chart Transformer (\our)\xspace}

\newcommand\blfootnote[1]{%
  \begingroup
  \renewcommand\thefootnote{}\footnote{#1}%
  \addtocounter{footnote}{-1}%
  \endgroup
}

\makeatletter
\DeclareRobustCommand\onedot{\futurelet\@let@token\@onedot}
\def\@onedot{\ifx\@let@token.\else.\null\fi\xspace}

\def\eg{\emph{e.g}\onedot} 
\def\ie{\emph{i.e}\onedot}

\def\etal{\emph{et al}\onedot}
\makeatother

\usepackage[capitalize]{cleveref}
\crefname{section}{Sec.}{Secs.}
\Crefname{section}{Section}{Sections}
\Crefname{table}{Table}{Tables}
\crefname{table}{Tab.}{Tabs.}

\usepackage[accsupp]{axessibility}  

\begin{document}
\pagestyle{headings}
\mainmatter
\def\ECCVSubNumber{5040}  

\title{Classification-Regression for Chart Comprehension} 

\titlerunning{Classification-Regression for Chart Comprehension}
%
\author{Matan Levy\inst{1}$^\dagger$ \and
Rami Ben-Ari\inst{2}\index{Ben-Ari, Rami} \and
Dani Lischinski\inst{1}}
\authorrunning{Levy et al.}
%
\institute{The Hebrew University of Jerusalem, Israel \and
OriginAI, Israel}
\maketitle

\begin{abstract}
   Chart question answering (CQA) is a task used for assessing chart comprehension, which is fundamentally different from understanding natural images. CQA requires analyzing the relationships between the textual and the visual components of a chart, in order to answer general questions or infer numerical values. Most existing CQA {\it datasets} and {\it models} are based on simplifying assumptions that often enable surpassing human performance. In this work, we address this outcome and propose a new model that jointly learns classification and regression. Our language-vision setup uses co-attention transformers to capture the complex real-world interactions between the question and the textual elements. We validate our design with extensive experiments on the realistic PlotQA dataset, outperforming previous approaches by a large margin, while showing competitive performance on FigureQA. Our model is particularly well suited for realistic questions with out-of-vocabulary answers that require regression.
\end{abstract}

\blfootnote{$^\dagger$Part of this research was conducted at IBM Research AI, Israel.}

\section{Introduction}
Figures and charts play a major role in modern communication, help to convey messages by curating data into an easily comprehensible visual form, highlighting the trends and outliers. 
However, despite tremendous practical importance, chart comprehension has received little attention in the computer vision community. Documents ubiquitously contain a variety of plots. Using computer vision to parse these visualizations can enable extraction of information that cannot be gleaned solely from a document’s text. Recently, with the rise of multimodal learning methods, \eg, \cite{yang2016stacked, vilbert, miech2019howto100m, clip, textvqa, changpinyo2021conceptual, visdial, schwartz2020factor}, interest in chart understanding has increased \cite{figureqa, plotqa, chartqa, chaudhry2019leafqa, dvqa, prefil}. 

\begin{figure}[ht]
	\begin{center}
		\includegraphics[width=0.7\columnwidth]{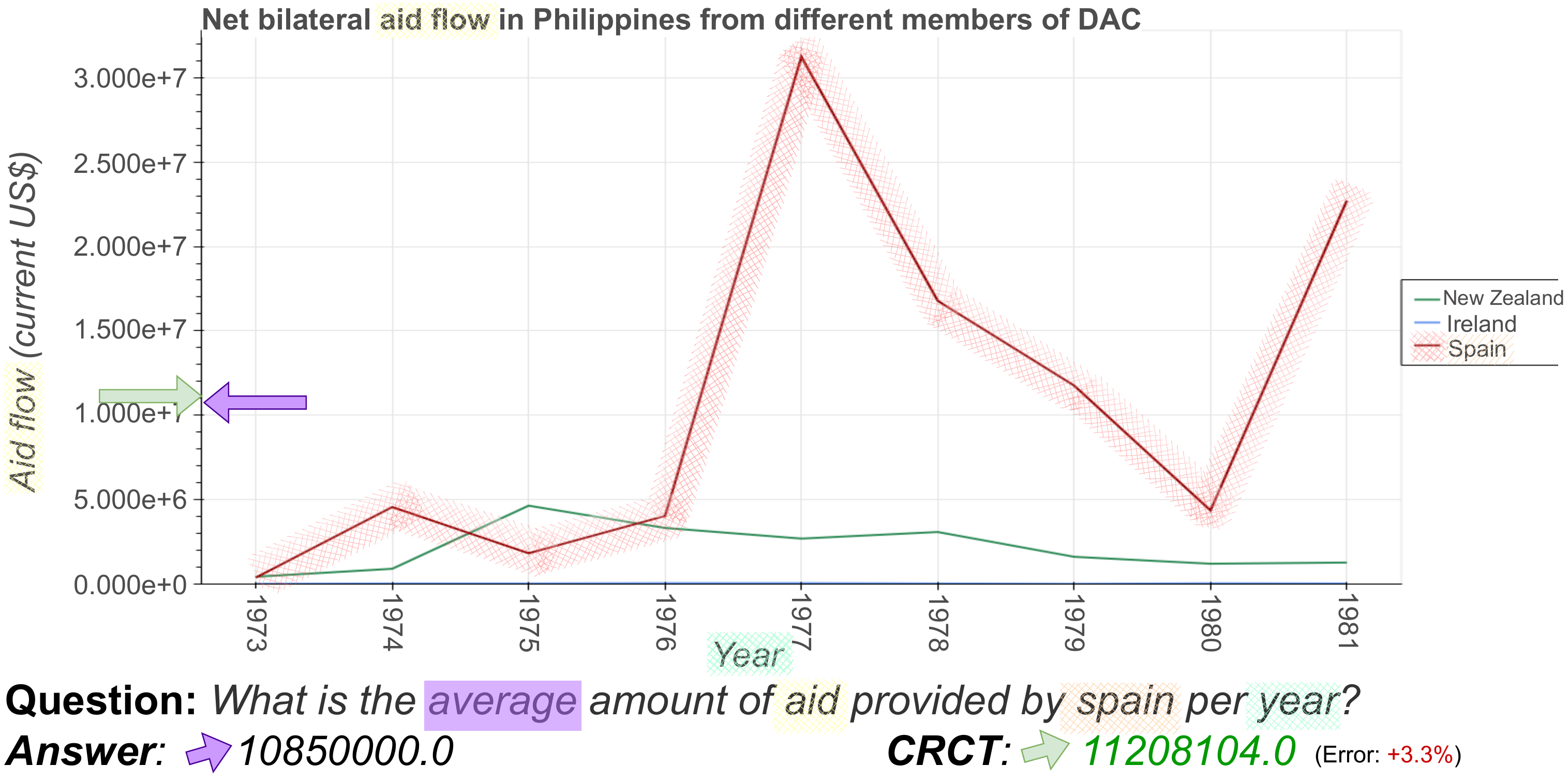}
	\end{center}
	\caption{Interactions marked on a sample from the PlotQA dataset~\cite{plotqa}, alongside with our \our prediction. We highlight the interacting parts/tokens with matching colors. Note the complexity of attention between the different modalities needed to correctly answer the question. The result predicted by \our and the ground truth answer are indicated by green and purple arrows.}
	\label{fig:Examples_for_Introduction}
\end{figure}

Studies on figure understanding (\eg, \cite{figureqa,plotqa}), commonly involve answering questions, a task known as Chart Question Answering (CQA). This task is closely related to Visual Question Answering (VQA), which is usually applied on natural images \cite{VQA, yang2016stacked, visdial, textvqa}. VQA is typically treated as a classification task, where the answer is a category, \eg, \cite{VQA, malinowski2015multiworld, acharya2018tallyqa, yang2016stacked}. In contrast, answering questions about charts often requires regression. Furthermore, a small local change in a natural image typically has limited effect on the visual recognition outcome, while in a chart, the impact might be extensive. Previous works have demonstrated that standard VQA methods perform poorly on CQA benchmarks~\cite{plotqa, dvqa}. A chart comprehension model must consider the interactions between the question and the various chart elements in order to provide correct answers. The complexity of such interactions is demonstrated in \cref{fig:Examples_for_Introduction}.
For example, failing to correctly associate a line with the correct legend text would yield an erroneous answer. 

Several previous CQA studies suggest a new dataset along with a new processing model, \eg, \cite{figureqa,dvqa,plotqa, chaudhry2019leafqa}. 
CQA datasets differ in several ways: (1) type and diversity of figures, (2) type and diversity of questions, (3) types of answers (\eg, discrete or continuous). While previous methods have recently reached a saturation level on some datasets, \eg, $94.9\%$ on FigureQA \cite{figureqa}, $92.2\%$ on LEAF-QA++ \cite{chartqa}, and $97.5\%$ on DVQA \cite{dvqa}, Methani \etal~\cite{plotqa} attribute this to the limitations of these datasets. Hence, they propose a new dataset (PlotQA-D), which is the largest and the most diverse dataset to date, with an order of magnitude more images/figures and $\times 4,000$ different answers. PlotQA-D further contains more challenging and realistic reasoning and data retrieval tasks, with a new model (PlotQA-M) achieving $22.5\%$ accuracy on this dataset, while human performance reached 80.47\% \cite{plotqa}.

In this paper we further explore the cause behind the saturation of various methods on previous data sets. We argue that similarly to early stages of VQA \cite{makingVinVQAmatter_CVPR2017}, several common datasets and benchmarks suffer from bias, over-simplicity and classification oriented Q\&A, allowing some methods to surpass human performance \cite{chartqa,prefil}. Next, we introduce a novel method called \ourFullname for CQA. We start with parsing the chart with a detector that extracts all of its textual and visual elements, which are then passed, along with the question text, to a dual branch transformer for bimodal learning. Our model features the following novelties: 1) In contrast to previous methods that encode only the question, our language model jointly processes all textual elements in the chart, allowing inter and intra relations between all textual and visual elements. 2) We show high generalization by dropping the common `string matching' practice (replacing question tokens with certain textual chart elements), and accommodating a co-transformer with pretrained BERT \cite{devlin2019bert}.  3) We introduce a new chart element representation learning, fusing multiple inputs from different domains. 4) Finally, a new hybrid prediction head is suggested, allowing unification of classification and regression into a single model. By jointly optimizing our model end-to-end for all types of questions, we further leverage the multi-task learning regime \cite{zhang2021survey}. 

We test our model on the challenging and more realistic dataset of PlotQA-D, as well as on FigureQA. Our results show that \our outperforms the previous method by a large margin on PlotQA-D (76.94\% vs.~53.96\% total accuracy), capable of matching previous results with 10\% of the training data.
We further analyze our model via explainability visualizations, revealing its limitations as well as strong capabilities.

\section{Related Work}
\label{sec:related}

In this section, we review existing CQA models, while focusing on the datasets in \cref{sec:Datasets}. In particular, we find that previous methods are often over-fitted to the type of datasets and corresponding questions/answers (Q\&A).

\begin{figure}[t]
\begin{center}
   \includegraphics[width=0.6\linewidth]{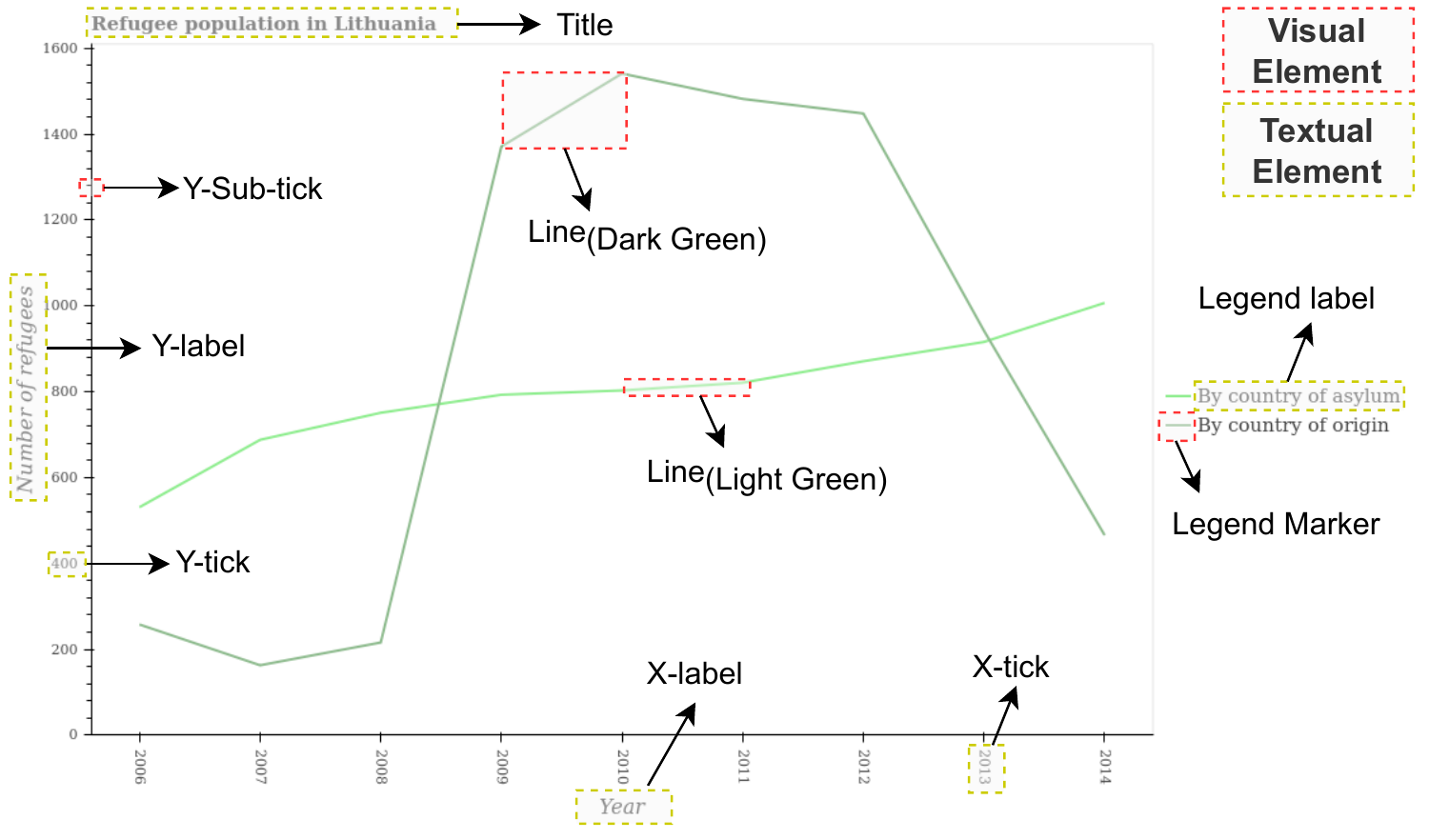}
\end{center}
   \caption{Examples of object annotations in train images.} 
\label{fig:visual_elements_names}
\end{figure}

Some CQA methods take the entire chart image as input to the model~\cite{figureqa, dvqa, prefil}, while others
first parse the image to extract visual elements using a detector \cite{plotqa, chartqa, chaudhry2019leafqa}. An example of chart elements and their corresponding class name, obtained from a detector, are shown in  \cref{fig:visual_elements_names}.

The pioneering model of Kahou \etal~\cite{figureqa} outputs binary (Yes/No) answers using a backbone pretrained on ImageNet fed into a Relation Network (RN) \cite{SantoroRBMPBL17}, in parallel to an LSTM \cite{lstm} used for question encoding. Removing the strong limitation to binary Q\&A, Kafle \etal \cite{dvqa} proposed a new dataset (DVQA) and a model referred to as SANDY. The dataset introduces new question types with \emph{out-of-vocabulary (OOV)} answers. These answers are chart specific (\eg, {\it Which item sold the most units in any store?}) and do not necessarily appear in the training set. The SANDY model is a classification network (SAN \cite{yang2016stacked}) with DYnamic encoding. In their approach, each text element in the chart is associated with a unique token in a dynamic encoding dictionary, based on the text location. These elements are then added to the dynamic list of answer classes. Kafle \etal\cite{prefil} later introduced PReFIL, another detector-free model with two branches: a visual branch based on DenseNet \cite{densenet}, and a text branch based on LSTM to encode the question. For bimodal fusion, they apply a series of 1$\times$1 convolutions on concatenated visual and question features.

Singh and Shekar \cite{chartqa} introduced STL-CQA, a new detector-based approach, combining transformers followed by co-transformers \cite{neuralMachineTranslation_ICLR15}.
Their method however, relies on replacement of tokens from the question with their string match in the chart, therefore tailored to the dataset question generator and is trained on its dictionary. As also claimed by the authors, STL-CQA is likely to fail in real cases where entities are addressed through their variations, which is the case in a reality as represented also in the PlotQA-D dataset.

All the above methods use only a classification head, without a regression capability, strongly limiting the generalization of these methods to realistic charts. OOV answers are therefore limited only to values appearing in the chart's image or {\it seen in train set} and added a-priori to the answer classes (see \cref{table:dataset_table}, \cref{sec:Datasets}). They commonly overlook the lingual relations between the chart's text, such as the relations between the content of the title, the legend, and the question. Instead, they only rely on the position of the text in the chart as a hint for its class.
Nevertheless, PReFIL showed overall accuracy above 93\% on FigureQA and DVQA surpassing human performance. Recent results shown in \cite{plotqa} imply that these datasets are strictly ``forgiving'' with respect to regression capability and lingual interactions between the questions and chart text (see \cref{sec:Datasets}).

Recently, Methani \etal \cite{plotqa} introduced a new method (PlotQA-M) and dataset (PlotQA-D). To the best of our knowledge, this is the first model to address the regression task, suggesting a solution for reasoning on realistic charts. PlotQA-M uses a visual detector and two separate pipelines. In a staging structure, a trained classifier switches between the pipelines, one handling fixed vocabulary classification, and the other for dealing with OOV and regression. In its OOV branch, PlotQA-M first converts the chart to a table and uses a standard table question-answering \cite{pasupat-liang-2015-compositional}, to generate an answer. This pipeline branching complicates the model requiring each pipeline to be   optimized separately and trained on a separate subset of the data, missing the impact of multi-task learning, which we further show as a strong advantage. Furthermore, PlotQA-M inter and intra visual-text interactions from the chart image are only determined through question encoding and a preprocessing stage using prior assumption on proximity between chart elements.
\section{Datasets}
\begin{table*}[t]
\caption{CQA datasets comparison. Real world vocabulary refers to axes variables. Some datasets apply question paraphrasing (par.)
}
\vspace{2mm}
\resizebox{\textwidth}{!}{%
\begin{tabular}{@{}lcccccccccccc@{}}
\toprule
\multicolumn{1}{c}{Dataset} & \begin{tabular}[c]{@{}c@{}}\#Plot\\ types\end{tabular} & \begin{tabular}[c]{@{}c@{}}\#Plot\\ images\end{tabular} & \begin{tabular}[c]{@{}c@{}}\#Q\&A\\ pairs\end{tabular} & \begin{tabular}[c]{@{}c@{}}Avg. question\\ length\end{tabular} & \begin{tabular}[c]{@{}c@{}}Q\&A\\ \#Templates\end{tabular} & \begin{tabular}[c]{@{}c@{}}\#Unique\\ answers\end{tabular} & \begin{tabular}[c]{@{}c@{}}Open\\ vocab.\end{tabular} & \begin{tabular}[c]{@{}c@{}}Real World\\ Vocabulary\end{tabular} & \multicolumn{1}{l}{\begin{tabular}[c]{@{}l@{}}Semantic\\ Relations\end{tabular}} & \begin{tabular}[c]{@{}c@{}}Bbox\\ Ann.\end{tabular} & \begin{tabular}[c]{@{}c@{}}Regression\\ answers\end{tabular} & \begin{tabular}[c]{@{}c@{}}Publicly\\ Available\end{tabular} \\ \midrule
FigureQA & 4 & 180k & 2.4M & 33.39 & \begin{tabular}[c]{@{}c@{}}15\\ (no variations)\end{tabular} & 2 & \xmark & \begin{tabular}[c]{@{}c@{}}\xmark\\ \small{(100 colors names)}\end{tabular} & \xmark & \cmark & \xmark & \cmark \\
DVQA & 1 & 300k & 3.5M & 55.22 & \begin{tabular}[c]{@{}c@{}}26\\ (w{\textbackslash}o par.)\end{tabular} & 1.5k & \begin{tabular}[c]{@{}c@{}}\cmark\\ \small{(Strings)}\end{tabular} & \begin{tabular}[c]{@{}c@{}}\xmark\\ \small{(1K nouns)}\end{tabular} & \xmark & Partial & \xmark & \cmark \\
LEAF-QA & 5 & 246k & 1.9M & - & \begin{tabular}[c]{@{}c@{}}35\\ (with par.)\end{tabular} & 12k & \begin{tabular}[c]{@{}c@{}}\cmark\\ \small{(Strings)}\end{tabular} & \cmark & \cmark & \cmark & \xmark & \xmark \\
LEAF-QA++ & 5 & 246k & 2.6M & 65.65 & \begin{tabular}[c]{@{}c@{}}75\\ (with par.)\end{tabular} & 25k & \begin{tabular}[c]{@{}c@{}}\cmark\\ \small{(Strings)}\end{tabular} & \cmark & \cmark & \cmark & \xmark & \xmark \\
PlotQA-D1 & 3 & 224k & 8.2M & 78.96 & \begin{tabular}[c]{@{}c@{}}74\\ (with par.)\end{tabular} & 1M & \begin{tabular}[c]{@{}c@{}}\cmark\\ \small{(Strings, Floats)}\end{tabular} & \cmark & \cmark & \cmark & \begin{tabular}[c]{@{}c@{}}\cmark\\ \small{(29.86\%)}\end{tabular} & \cmark \\
PlotQA-D2 & 3 & 224k & 29M & 105.18 & \begin{tabular}[c]{@{}c@{}}74\\ (with par.)\end{tabular} & 5.7M & \begin{tabular}[c]{@{}c@{}}\cmark\\ \small{(Strings, Floats)}\end{tabular} & \cmark & \cmark & \cmark & \begin{tabular}[c]{@{}c@{}}\cmark\\ \small{(88.84\%)}\end{tabular} & \cmark \\ \bottomrule
\end{tabular}
}

\label{table:dataset_table}
\end{table*}

\label{sec:Datasets}
In this section we discuss the properties of existing CQA datasets, emphasizing the bias they introduce into the models and the evaluation methodologies that were proposed. \cref{table:dataset_table} presents various properties of these datasets that may strongly impact the realism and generalization of the results to a real world application. This is an extended version of a table shown by Methani \etal \cite{plotqa}.

Probably the most popular CQA datasets/benchmarks are FigureQA \cite{figureqa} and DVQA \cite{dvqa}, both of which are publicly available. FigureQA consists of line plots, bar charts, pie plots, and dot line plots, with question templates that require binary answers. The plot titles and the axes label strings are constant; the axes range is mostly in $[0, 100]$ with low variation; and the legends are chosen from a small set of {\it color names} (see example in supplementary material). These properties detract from the realism of this dataset.

DVQA \cite{dvqa} contains a single type of charts (bar charts), but offers more complexity in Q\&A. The answers are no longer only binary, and may be out of vocabulary (OOV). Questions are split to three conceptual types: {\bf Structural}, {\bf Data retrieval} and {\bf Reasoning}.
Structural questions refer to the chart's structure (\eg, \emph{How many bars are there?}). Data retrieval questions require the retrieval of information from the chart (\eg, \emph{What is the label of the third bar from the bottom?}). Reasoning questions demand a higher level of perceptual understanding from the chart and require a combination of several sub-tasks (\eg, \emph{Which algorithm has the lowest accuracy across all datasets?}). Yet, this dataset suffers from lack of semantic relations between the text elements (\eg, bar and legend labels are randomly selected words), and the range of values on the Y-axis is limited. About 46 out of 1.5K unique answers are numeric, consisting of integers with the same values in the train and test sets, allowing a classification head to handle data retrieval and reasoning.

Two more datasets LEAF-QA \cite{chaudhry2019leafqa} and LEAF-QA++ \cite{chartqa}, have fewer Q\&A pairs than DVQA, but several types of charts, and use a real world vocabulary with semantic relations (see \cref{table:dataset_table}).
However, they are both proprietary. All the mentioned datasets share a strong limitation, lack of regression Q\&A, indicated by their question templates and their discrete answer set. PlotQA-D \cite{plotqa} is, however, the largest and most comprehensive publicly released dataset to date. This dataset consists of charts generated from real-world data, thereby exhibiting realistic lingual relations between textual elements. The questions and answers are based on multiple crowd-sourced templates. PlotQA-D consists of three different chart types: line-plots, bar-charts (horizontal and vertical), and dot line plots. The range of the Y-axis values is orders of magnitudes larger (up to $[0,3.5 \times 10^{15}]$) with non-integer answers generally not seen in training, resulting over 5.7M of different answers. In contrast to previous datasets, PlotQA-D often requires a regressor for correctly answering questions. Nearly 30\% and 90\% of questions require regression in PlotQA-D1 and PlotQA-D2 respectively (see \cref{table:dataset_table}). To the best of our knowledge, PlotQA-D is currently the most realistic publicly available dataset.
PlotQA-D offers two benchmarks, the first version of the dataset PlotQA-D1, and its extended version PlotQA-D2, which contains the former as a subset (28\% of the Q\&A pairs on the charts). The majority of PlotQA-D2 question types require regression (see the suppl.~material). 
We believe that saturated performance on DVQA (97.5\%), probably attributed to a single plot type and having only 1.5K unique in contrast to 5.7M answers in PlotQA-D, makes it inappropriate for regression benchmarking.
\section{Method}%
\begin{figure*}[t]
\begin{center}
   \includegraphics[width=1\linewidth]{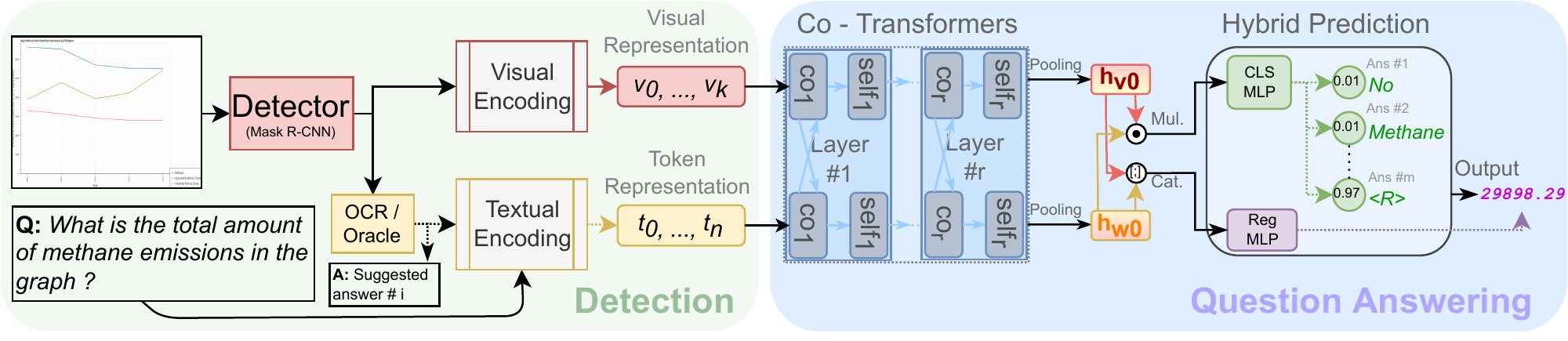}
\end{center}
   \caption{Our \ourFullname  network architecture consists of two stages of detection and question answering. The detection stage (left) provides bounding boxes and object representations of the visual and textual elements (see Fig. \ref{fig:visual_elements_names}). These features, along with the question text, enable the co-transformers in the second stage (right) to fuse both visual and textual information into a pooled tuple of two single feature vectors $\{{\bf h_{v_0}},{\bf h_{w_0}}\}$. Next, our hybrid prediction head containing two different MLPs, outputs a classification score and a regression result. co$_i$/self$_i$: co/self attention.
   }
\label{fig:all_arc}
\end{figure*}

We present an overview of our \our architecture for CQA in \cref{fig:all_arc}. In our approach, the image is first parsed by a trained object detector (see object classes in \cref{fig:visual_elements_names}). The output of the parsing stage are object classes, positions (bounding boxes), and visual features. All of the above are projected into a single representation per visual element, then stacked to form the \emph{visual sequence}. Similarly, each textual element is represented by fusing its text tokens, positional encoding and class. Together with the question text tokens, we obtain the \emph{text sequence}.
The two sequences are fed in parallel to a bimodal co-attention-transformer (co-transformer). The output of the co-transformer are pooled visual and textual representations that are then fused by Hadamard product and concatenation, and fed into our unified classification-regression head. In the next sections we describe the train and test configurations in detail.

\begin{figure}[t]%
    \centering
    \subfloat[\centering Visual Representation.]{{\includegraphics[width=0.49\linewidth]{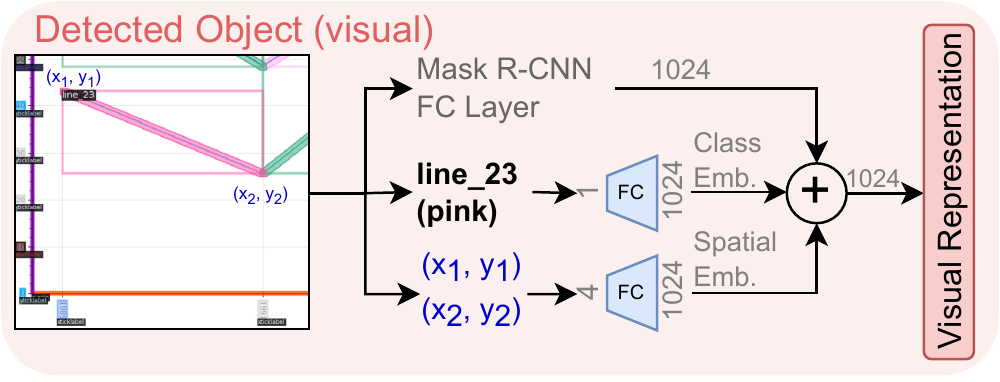} }\label{fig:feature_embeddings_vis}}%
    \hfill
    \subfloat[\centering Textual Representation (per token).]{{\includegraphics[width=0.49\linewidth]{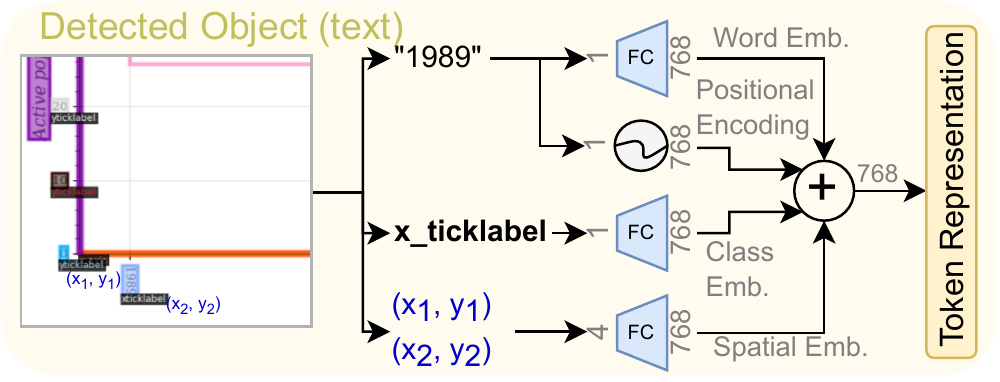} }\label{fig:feature_embeddings_txt}}%
    \caption{Chart element representations. The relevant information for representing each type of element is summed into a single vector.}
    \label{fig:feature_embeddings}
\end{figure}

{\bf Visual Encoding:} The visual branch encodes all the visual elements in the chart, \eg, line segments or legend markers. For visual encoding we train a Mask-RCNN \cite{he2017mask} with a ResNet-50 \cite{he2015deep} backbone. Object representations are then extracted from the penultimate layer in the classification branch. In our detection scheme objects are textual elements (\eg, title, xlabel) as well as visual elements (\eg, plot segment) as shown in \cref{fig:visual_elements_names}.
We create a single representation per visual element by a learnable block as shown in \cref{fig:feature_embeddings_vis}. This block takes as input the 4D vector describing the bounding box (normalized top-left and bottom-right coordinates), the class label and the object representation produced by the detector (encapsulating \eg, the line direction), and projects them to an embedding space (1024D).

Object colors are generally encoded in the representation output from the detector. However the actual colors are often important for linking the legend marker to the legend label (text), allowing the connection between the question and the target line or bar in the chart. Our observation shows that training the detector with decomposition of graphs to colors, boosts the performance.
Finally, our visual element representations form a sequence, is denoted by ${v_1, ..., v_k}$.
We further add the global plot representation ($v_0$) as [CLS] token.

{\bf Text Encoding:} Raw text is handled with a pretrained BERT \cite{devlin2019bert}. The textual features are derived from the question and the text contained within the chart, such as the axes labels, legends and title. In contrast to VQA where the lingual part includes only the question, in CQA there are additional text elements that are essential for chart comprehension. Text position in the chart carries important information.
In this study, we encode the textual elements in a concatenated version, separated with the special [SEP] token, followed by the question and an answer with the special token [CLS] on top ($t_0$). In contrast to previous work \cite{vilbert,chartqa, dvqa, figureqa, yang2016stacked}, where only the question (or question + answer) was encoded, here the text encoder is generalized to include all textual elements enriched with their spatial location and class. This approach allows free data-driven interaction between different visual and textual elements, \eg, the legend marker and its corresponding text, as well as interactions between text sub-elements, \eg, the answer and {\it part} of the Y-axis label or title. To this end, we create a new representation from all the textual elements in the chart by fusing the word embedding, the positional encoding, the text location in the chart and the text class embedding. This fusion is carried out through a MLP layer, including projection and summation as shown in \cref{fig:feature_embeddings_txt}. 

\subsection{Associating Visual and Textual Elements}
\label{sec:associating_visual_and_textual}
For multi-modal interaction we rely on the co-attention architecture that was first suggested for machine translation in \cite{neuralMachineTranslation_ICLR15}. 
This model contains two different sequence to sequence branches: visual and textual, as shown in \cref{fig:all_arc}. The information in the two streams is fused through a set of attention block exchanges, called co-attention. We use a transformer with 6 blocks of two encoders with {\it co-} and {\it self-} attention. Each encoder computes a query $Q$, key $K$, and value $V$ matrices, followed by feed-forward layer, skip connections and normalization \cite{vaswani2017attention}. In order to exchange the information between the modalities, the {\it co-}transformer's keys and values at each stream are mutually exchanged resulting a cross-modality attention.
Finally, the resulting ${\bf \{h_{v_0}},{\bf h_{w_0}}\}$ pooling tokens (indicated by [CLS] special token) are forwarded to the classification and regression heads (see \cref{fig:all_arc}). For more details, see suppl.~material.

\subsection{Question Answering Stage}
\label{sec:question_answering_stage}
Similar to previous work \cite{chartqa, prefil, chaudhry2019leafqa, dvqa, plotqa} and in order to allow fair comparison, we use an oracle to recognize the extracted text elements. The oracle is a perfect text recognition machine, and is used to disentangle the impact of OCR accuracy. Previous work frequently assume a perfect text detector, \eg, \cite{figureqa, dvqa, chartqa, plotqa}. In this work however, we explicitly account for inaccuracies in the detector by considering only text elements from the oracle with $\textit{IoU}>0.5$. We then create the set of possible answers for classification, composed of {\it in-vocabulary} (\eg, Yes / No) and {\it out-of-vocabulary} (OOV) answers (\eg, the title or specific legend label).
OOV additional classes (dynamically added) allow dealing with chart specific answers that has not been seen during training. To predict the correct answer, we train the model with binary cross-entropy loss. To this end, we concatenate the answer to the question in the textual branch, pass it through the model and evaluate a score in $[0,1]$ range (see \cref{fig:all_arc}). This score indicates the model's certainty whether the answer is aligned with the question (correct) or not (wrong).

\subsection{Unified Prediction}
\label{sec:hybrid_prediction}
Previous works frequently use only a classification head, overlooking regression \cite{figureqa,dvqa,chaudhry2019leafqa,chartqa}, or use a totally separate pipeline for the regression task \cite{plotqa}. In classification based methods, the answers are restricted to discrete values, that are part of the numeric values appearing on the chart. This approach strongly limits the generalization, lacking the capability to predict unseen numeric values or charts with unseen ranges. 
In this work, we propose a novel hybrid prediction head allowing unified classification-regression. To this end, we add a regression soft decision flag $\langle R \rangle$ as an answer class, followed by a regressor. During training the model learns which type of questions require regression by choosing the $\langle R \rangle$ class as the correct answer. A separate and consequent regression is then applied to generate the answer (see \cref{fig:all_arc}). Note that during training, the loss changes dynamically from BCE loss for classification and L1 loss for regression, so the network is jointly optimized for classification and regression. During train, we vanish the regression loss when the correct class is not $\langle R \rangle$. The hybrid prediction allows joint training on all types of Q\&As, leveraging multi-task learning. 

\subsection{Implementation Details}
For training the \our we use two stages. We first train a Mask-RCNN \cite{he2017mask} from which the visual features are derived, using Detectron2 \cite{wu2019detectron2} library. We then train the co-transformer model for 20 epochs with linear learning rate scheduler. We use binary cross entropy loss for the classification component and L1 loss for regression. For answer alignment prediction (as described in \cref{sec:question_answering_stage}), we generate negative examples by randomly assigning wrong answers to questions. Training our model on PlotQA-D1 took 3.5 days on two Nvidia RTX-6000 GPUs. The inference computational cost is proportional to the size of candidate answers. In our experiments the inference time took 0.23 seconds per question. Our code and models are publicly available at \href{https://github.com/levymsn/CQA-CRCT}{https://github.com/levymsn/CQA-CRCT}.


\section{Evaluation}
\label{sec:Evaluation}
As evaluation benchmark we opted for PlotQA-D and FigureQA datasets, being fully annotated to train a detector (DVQA lacks the important annotation of legend markers). Yet, we focus our analysis on PlotQA-D for several reasons: (1) Publicly available to allow benchmarking. (2) The scale: Over $\times$10 larger Q\&A pairs and over $\times$1000 more unique answers, than the predecessors (see \cref{table:dataset_table}); (3) Highly variable axis scale; (4) Having diverse and realistic questions/answers with rich vocabulary titles, legend labels, X and Y labels including initials gathered from real figures; (5) Most importantly, question types that require regression and therefore reflect a realistic case for CQA.

In terms of methods to compare with, we searched for publicly available code or assessments on the chosen datasets. 
To allow a fair comparison to previous methods, in addition to PlotQA-M,  we further test PReFIL~\cite{prefil} on PlotQA-D. To this end, we trained PReFIL on PlotQA-D1. We chose PreFIL due to it's high performance on DVQA and FigureQA and as a representative candidate for previous methods that rely on classification and lack a regression capability. Since PReFIL has only a classification head we quantized the numeric values into Y-ticks and added them to the dynamic classification head in training and also at test (a common practice, also performed in PReFIL \cite{prefil}). For sake of analysis and to allow a fair comparison we show the PReFIL results for numeric evaluation with various error tolerances (see \cref{fig:crct_prefil_comp}).

To handle the wide range of Y-axis values in PlotQA-D, we normalize values to $[-1, 1]$ (by detecting X-Y axes and their values). This improves convergence and enables scale invariant prediction. We output answers in the same range. 
\subsection{Results}%
We train our model on PlotQA-D1 dataset, that consists of one third of PlotQA-D2 in questions, while testing on both PlotQA-D1 and PlotQA-D2 test sets. We show significant improvements on both test sets. Results are shown as average accuracy over the test set and accuracy breakdown per-question category.

{\bf Comparison to previous methods}: 
\cref{tab:v1_accuracies} summarizes the results on PlotQA-D1 test set. In general, we outperform PlotQA-M in all categories by a large margin. For instance, the gaps for Data Retrieval and Reasoning are 48.8\% (94.52\% vs. 45.68\%) and 23.7\% (54.87\% vs. 31.20\%) {\it absolute} points, respectively. Finally, on average we achieve 76.94\% accuracy, compared to 53.96\% of PlotQA-M.
While outperforming PlotQA-M when trained on the same train set, in
the next experiment we show the extent of train data reduction that can be allowed to match the previous results of PlotQA-M. This experiment shows that as little as 10\% of training data (randomly selected) are
already sufficient to reach this goal. (see \our-10\% in \ref{tab:v1_accuracies}).

With respect to PReFIL, while we show comparable results on the Structural question category, containing classification type questions, \our is superior to PReFIL in all other categories. As expected, PReFIL performs poorly on Data Retrieval and particularly Reasoning Q\&As (only 31.66\% vs 54.87\% for our CRCT) due to lack of regression capability.
In total average accuracy we surpass both PlotQA-M and PReFIL by 23\% and 19\% \emph{absolute} points, respectively. Interestingly, with our quantization scheme training of PReFIL, it outperforms PlotQA-M, in all categories. 
\begin{table}[b]
\centering
\caption{Accuracies [\%] on PlotQA test sets. Values in each column indicate average accuracy per-question category. \our and PReFIL are trained on the PlotQA-D1 subset. PReFIL results are reproduced. `\our-10\%' indicates our results with training on 10\% of the PlotQA-D1 train set. S, D and R stand for Structural, Data Retrieval and Reasoning question categories, respectively}
\label{table:accuracies_v1_v2}
\subfloat[\centering Evaluation on PlotQA-D1 test set]
{
\resizebox{0.47\columnwidth}{!}{%
\huge
\begin{tabular}{lcccc}
\hline
Method & ~S~ & ~D~ & ~R~ & ~Overall \\ \hline
PlotQA-M\cite{plotqa} & 86.31 & 45.68 & 31.2 & 53.96 \\
\our-10\% & 87.15 & 74.71 & 29.19 & 57.75 \\
PReFIL\cite{prefil} & \textbf{96.66} & 58.69 & 31.66 & 57.91 \\
\our (ours) & 96.13 & \textbf{94.52} & \textbf{54.87} & \textbf{76.94} \\ \hline
\end{tabular}

}
\label{tab:v1_accuracies}
}
\hfill
\subfloat[\centering Evaluation on PlotQA-D2 test set]{
\resizebox{0.47\columnwidth}{!}{%
\huge
\begin{tabular}{lcccc}
\hline
Method & ~S~ & ~D~ & ~R~ & ~Overall \\ \hline
PReFIL\cite{prefil} & \textbf{96.66} & 21.9 & 3.9 & 10.37 \\
PlotQA-M\cite{plotqa} & 75.99 & 58.94 & 15.77 & 22.52 \\
\our (ours) & 96.23 & \textbf{66.65} & \textbf{25.81} & \textbf{34.44} \\ \hline
\end{tabular}
}
\label{tab:v2_accuracies}
}
\end{table}

Due to extreme computational demand for train on PlotQA-D2, in the next experiment we train PReFIL and CRCT on PlotQA-D1 train set and report the results on PlotQA-D2 test set in \cref{tab:v2_accuracies}. Note that for PlotQA-M we report the result from \cite{plotqa} with the model trained on whole PlotQA-D2. 
These results show that even when we train on PlotQA-D1 dataset we are able to outperform PlotQA-M trained on $\times$3 larger size data, in all categories, often with significant margin. Our CRCT is superior here also to PReFIL with average accuracy of 34.44\% vs. 10.37\%. Note the poor performance of PReFIL on Reasoning category, from which many questions require regression, reaching 3.9\% comparing 25.81\% in CRCT. These results show the significance of our hybrid classification-regression capability.

{\bf Regression Performance}: 
The accuracy of regression errors are often measured by $L_2$ or $L_1$ differences or by ER-error rate. 
In PlotQA-D \cite{plotqa}, a regression answer is considered correct if it falls within $\pm5\%$ tolerance from the ground truth value.
This measure, however, is proportional to the true value, vanishing (no tolerance) for true values near zero. We therefore suggest the \emph{tick-based} error measure as more appropriate for extraction of numerical values. To this end we suggest a constant gap per-chart, defined as a fraction of units between two consecutive sub-ticks (see \cref{fig:visual_elements_names}) \eg, $\nicefrac{1}{4}$ sub-tick.

\begin{figure}[ht]%
    \centering
    \subfloat[\centering {The prevalence of CRCT's answers that fall in certain error range.}]{{\includegraphics[width=0.49\linewidth]{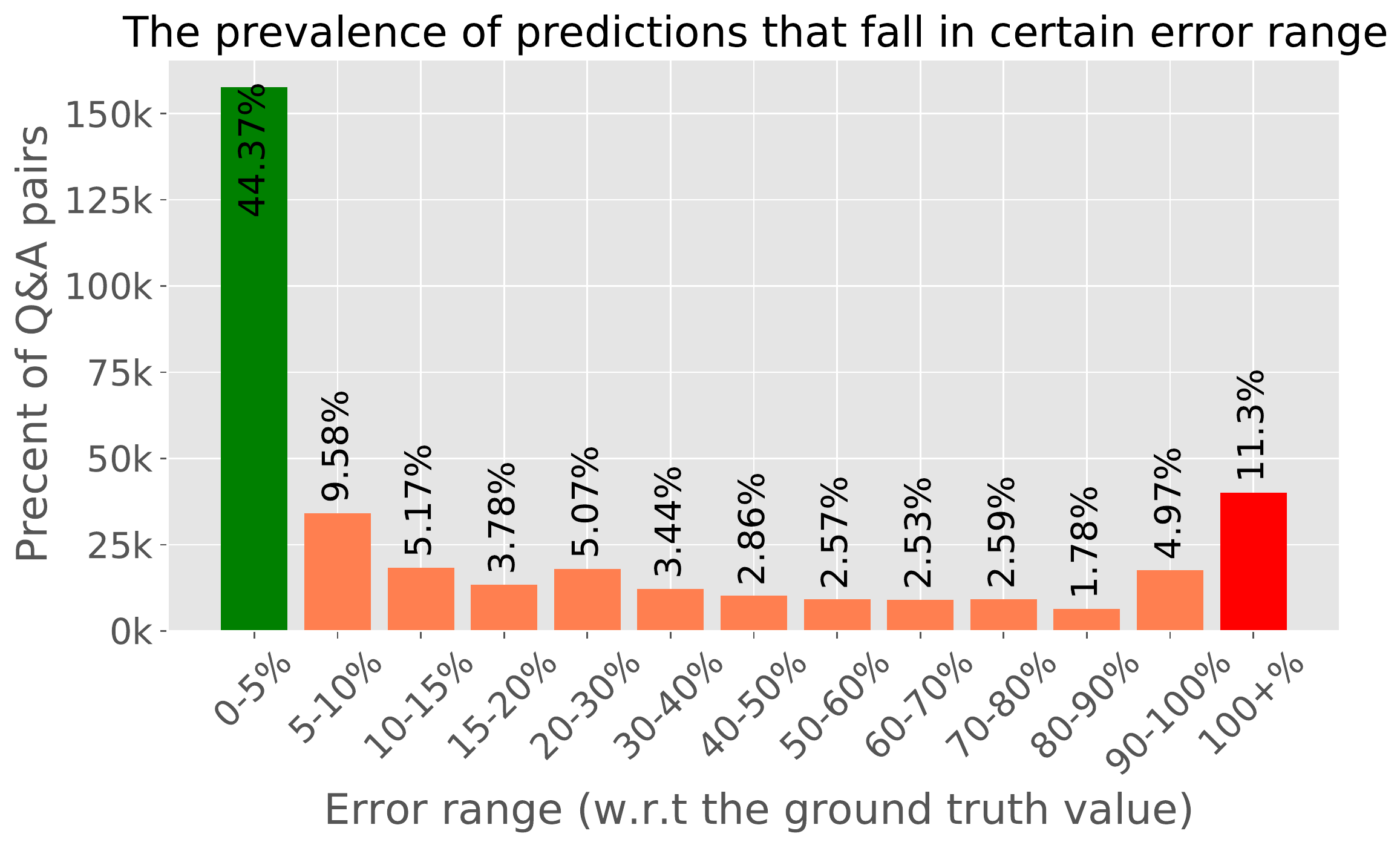} }\label{fig:regressor_performance}}%
    \hfill
    \subfloat[\centering Accuracy for different sub-tick error range (tolerance). ]{{\includegraphics[width=0.49\linewidth]{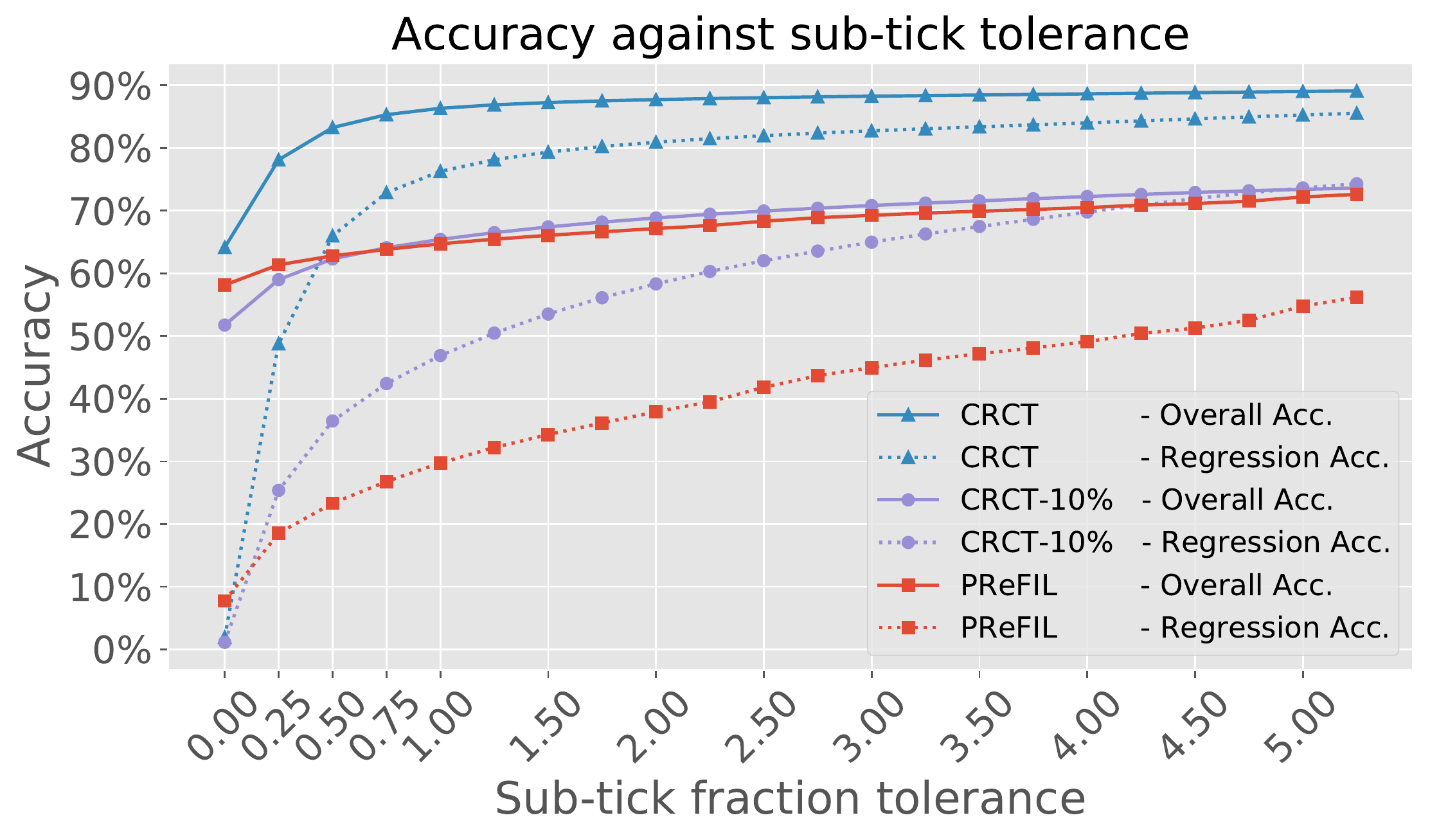}}\label{fig:crct_prefil_comp}}
    \caption{Model regressor performances on PlotQA-D1. In \ref{fig:regressor_performance}, the green column shows the ``correct" answers \ie fall in 5\% tolerance. 11.3\% of the answers (red) miss the target by more than 100\%. In \ref{fig:crct_prefil_comp}, $x=0$ indicates exact match between prediction and ground truth (zero tolerance).}
\end{figure}

In PlotQA-D1, $29\%$ of the questions require regression. Following PlotQA \cite{plotqa}, we show in \cref{fig:regressor_performance} CRCT accuracy distribution considering the error rate (ER) measure.
We observe that 44.37\% of the answers are within $\pm 5\%$ of the true value. The prevalence of errors decreases in higher tolerance ranges except the {\it outlier} in the tail, indicating that 11.3\% of the answers were over 100\% off the true value. As expected, we observe that CRCT error distribution indeed accumulates near zero true values (see suppl. material), justifying the advantage of value invariant error measure. \cref{fig:crct_prefil_comp} shows the variation of regression accuracy with increased tolerance (as sub-tick fraction) for CRCT and PReFIL. CRCT achieves over 85\% total accuracy and 78\% regression accuracy for 1 sub-tick tolerance. Note the large gap w.r.t PReFIL through all the range as well as the drop in CRCT-10\% that obtained similar accuracy to PlotQA-M (\cref{fig:regressor_performance}).
For visual examples of our CRCT model on regression assignment see Figures \ref{fig:Examples_for_Introduction}, \ref{fig:visualization_639414} and the suppl.~material.

{\bf Results on FigureQA: } Although our model's strength is in general Q\&A with regression, we also test our model on the binary answer data set of FigureQA \cite{figureqa}. FigureQA's training set was generated using different 100 colors. This dataset contains two families of validation and test sets. The first family is the Val-1/Test-1 sets, that was generated using the original color schemes as in the train set. On the contrary, Val-2/Test-2 sets consist of alternate color scheme that {\it was not seen} in the train set at all.
\cref{table:figureqa_results} presents a comparison on FigureQA dataset. \our shows comparable performance to SoTA on the original color scheme. While we outperform previous methods on the alternate color scheme sets, we reach an inferior performance w.r.t PReFIL. This test indicates a color sensitivity for our detector-based approach as we discuss in \cref{sec:summary}.
\begin{table}[t]
\centering
\caption{Accuracy on FigureQA dataset \cite{figureqa}. Second place is coloured in brown}
\vspace{-2mm}
\subfloat[\centering Original color scheme]
{
\resizebox{0.45\columnwidth}{!}{%
\begin{tabular}{p{0.22\textwidth}p{0.1\textwidth}p{0.1\textwidth}}
\hline
Model / Acc.                      & \multicolumn{1}{l}{Val.} & \multicolumn{1}{l}{Test} \\ \hline
RN\cite{figureqa}                 & -                        & 76.52                     \\
LEAF-Net\cite{chaudhry2019leafqa} & -                        & -                         \\
Zou et al. \cite{Zou2020AnAR}     & 85.48                    & 85.37                     \\
\our (ours)                       & \bf{\textcolor{Brown}{94.61}}                     & \bf{\textcolor{Brown}{94.23}}                     \\
PReFIL\cite{prefil}               & {\textbf{94.84}}           & {\textbf{94.88}}            \\ \hline
\end{tabular}
\label{table:figureqa_results_set_1}
}
}
\hfill
\subfloat[\centering Alternate color scheme]{
\resizebox{0.45\columnwidth}{!}{%
\begin{tabular}{p{0.22\textwidth}p{0.1\textwidth}p{0.1\textwidth}}
\hline
Model / Acc.                      & \multicolumn{1}{l}{Val.} & \multicolumn{1}{l}{Test} \\ \hline
RN\cite{figureqa}                 & 72.54                    & 72.40                     \\
LEAF-Net\cite{chaudhry2019leafqa} & 81.15                    & -                         \\
Zou et al. \cite{Zou2020AnAR}     & 82.95                    & 83.05                     \\
\our (ours)                       & \bf{\textcolor{Brown}{85.04}}                    & \bf{\textcolor{Brown}{84.77}}                    \\
PReFIL\cite{prefil}               & {\textbf{93.26}}           & {\textbf{93.16}}            \\ \hline
\end{tabular}
\label{table:figureqa_results_set_2}
}
}
\label{table:figureqa_results}
\end{table}

\begin{table}[t]
\small
\begin{center}
\caption{Ablation study with different configurations (see also Fig. \ref{fig:feature_embeddings}). All models are trained on 10\% of PlotQA-D1 train set, and evaluated on the entire PlotQA-D1 test set. S, D and R stand for Structural, Data Retrieval and Reasoning, respectively}
\vspace{2mm}
\resizebox{0.9\textwidth}{!}{%
\begin{tabular}{@{}lcccccc@{}}
\toprule
Method & Regression & Classification & ~~~~S~~~~ & ~~~~D~~~~ & ~~~~R~~~~ & Overall \\ \midrule
w/o Legend Marker & 14.76 & 65.02 & 81.13 & 56.01 & 27.05 & 50.45 \\
w/o {\it Textual} Class Emb. & 15.51 & 66.86 & 81.75 & 61.73 & 26.96 & 51.98 \\
w/o {\it Visual} Class Emb. & 17.35 & 73.68 & 85.06 & 73.09 & 30.57 & 57.36 \\
Only Bbox for{\it Visual} Feats. & 18.66 & 68.68 & 84.97 & 72.94 & 23.75 & 54.19 \\
Two Pipelines & 14.80 & 70.19 & 84.49 & 68.65 & 25.16 & 53.64 \\
CRCT & 20.74 & 72.86 & 87.15 & 74.71 & 29.19 & \textbf{57.75} \\ \bottomrule
\end{tabular}
\label{table:Ablation_Study}
}
\end{center}
\vspace{-8mm}
\end{table}

\section{Ablation Study}
 \cref{table:Ablation_Study} shows an ablation study of our method using different configurations. First we examine the impact of the {\it legend marker} (see \cref{fig:visual_elements_names}) as key element. Removing it from the input in the visual branch prevents the model to associate the question to the specific plots/bar in multi-graph chart. The results show drop in performance in all categories with total accuracy dropping from 57.75\% to 50.45\%. In the next two tests we show the impact of representation architecture on the end results. To this end we remove the class label embeddings from the visual and textual representation (\eg, `line\_23' or `x\_ticklabel' in \cref{fig:feature_embeddings}). Although noisy, these inputs derived from the detector, positively impact the results. Removing them, causes regression accuracy to drop from 20.74\% to 17.35\%, for visual and 15.51\% for text. We observe the best classification performance is achieved without the visual class embedding. However, this embedding is just one component of the visual representation (see \cref{fig:feature_embeddings_vis} - Class-Emb). In some cases Class-Emb is redundant to the visual representation, and removing it can slightly improve certain classification Q\&As, resulting in this outcome (\eg, where only textual elements are addressed). However, as \cref{table:Ablation_Study} shows, the slight improvement in classification task ($\sim1\%$) is traded with large degradation in regression accuracy ($\sim3\%$), resulting a lower total accuracy.
 When removing all features except the bounding box coordinates, from the visual representation, the total accuracy drops by 3.6\%. This shows the importance of all elements in our chart element representation model (see Fig. \ref{fig:feature_embeddings}). Finally, we examine the importance of the multi-tasking regime inherent in our unified classification-regression network. To this end we train our classification and regression network separately (similar to \cite{plotqa}). Assuming an oracle for routing classification and regression type questions to the proper network, we report the outcome accuracies. We observe performance drop on all categories emphasizing the importance of combining both regression and classification in \our's learning process. Our detector achieved AP50=0.90. Testing our model with ground truth detections had a negligible effect on the accuracy.

\section{Explainability} 
\begin{figure}[ht]
    \centering
    \subfloat[\centering Q: {\small\it In which year was the use of IMF credit in DoD minimum?}\newline GT: 1989, \our: \textcolor{ForestGreen}{1989}.]{{\includegraphics[width=0.49\linewidth]{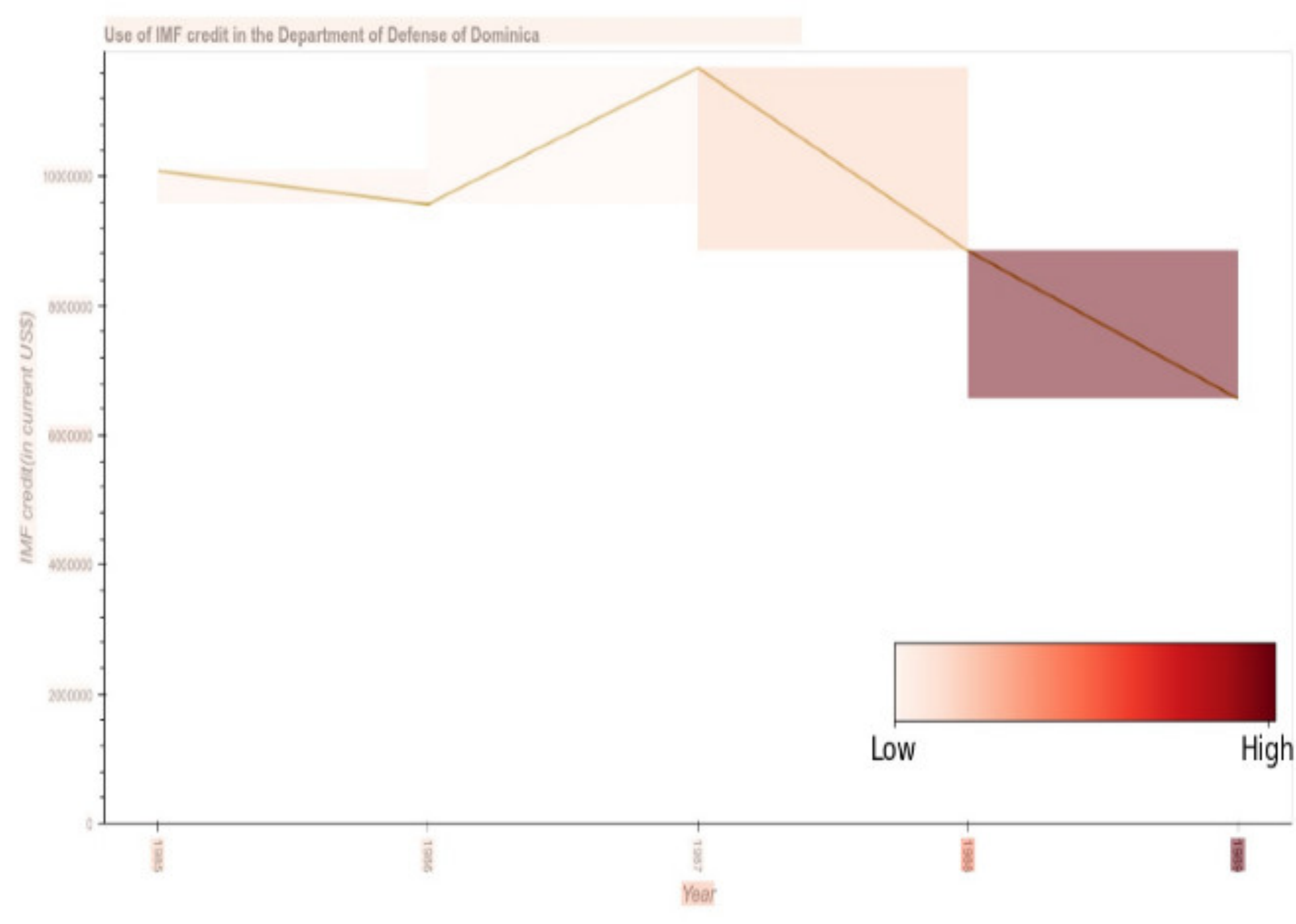} }\label{fig:visualization_639418}}%
    \hfill
    \subfloat[\centering Q: {\small\it What is the average percentage of labor force who received secondary education per country?}\newline GT: 48.05, \our: \textcolor{ForestGreen}{47.91} (Error: \textcolor{Red}{-0.29\%}).]{{\includegraphics[width=0.49\linewidth]{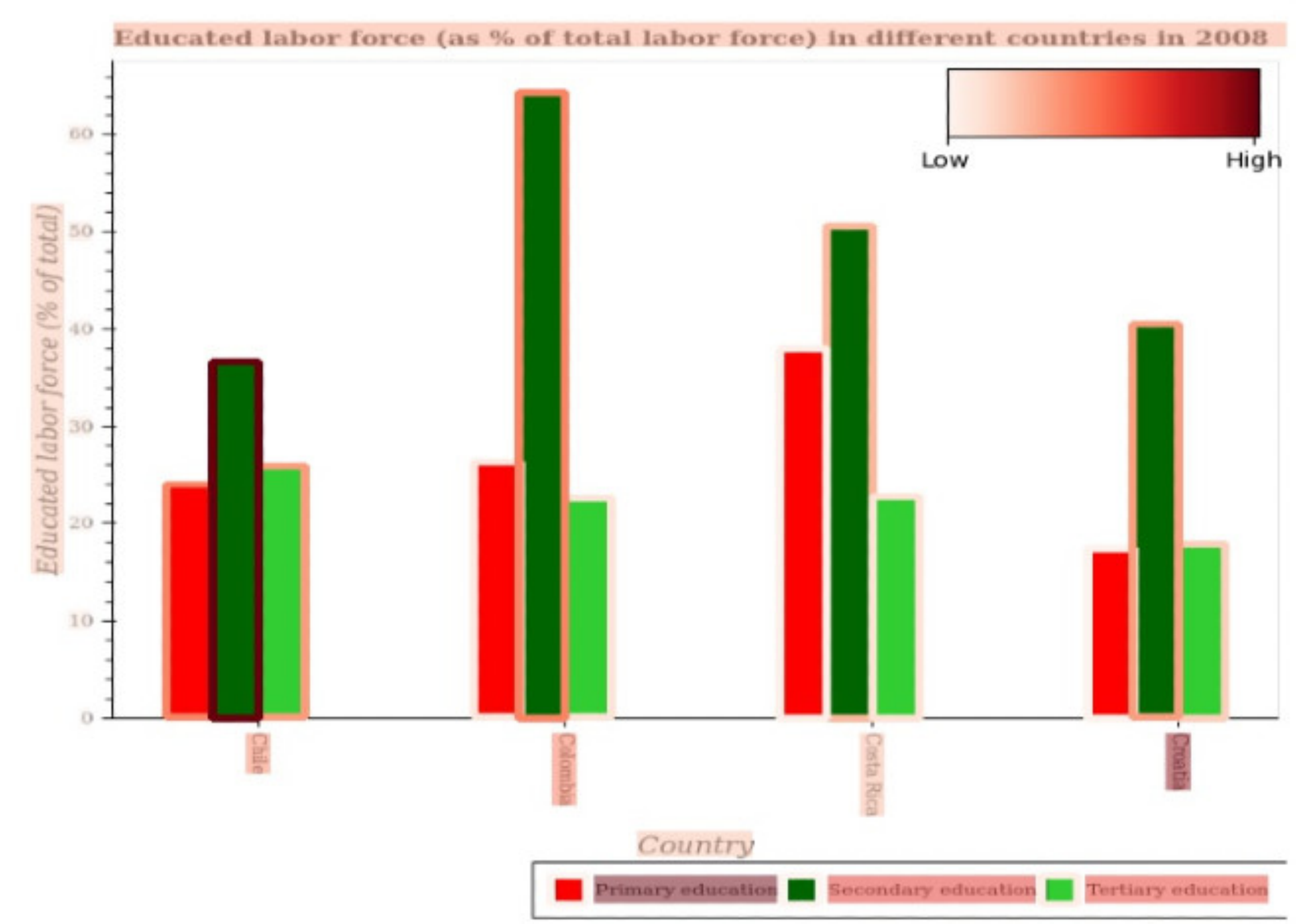} }\label{fig:visualization_639414}}%
    \caption{Test set visualizations. Warmer box color means higher influence.}
    \label{fig:visualizations}
\end{figure}

We provide visualizations for \our attention using the {\it Captum} package \cite{leino2018influencedirected, captum2019github}. 
Often, relatively few units in a NN are highly influential towards a particular class \cite{leino2018influencedirected}. Considering the true answer, we integrate over the input gradients to find the most influential features. We then color code the image to indicate the regions in the chart, visual or textual, that the network found influential in answering the posed question. \cref{fig:visualizations} shows such visualization maps over charts, on examples from the test set.  In \cref{fig:visualization_639418} \our correctly ``looks'' at the x-tick at the global minimum in the plot and on the corresponding x-label, when asked about the \emph{minimum} argument. Fig. \ref{fig:visualization_639414} shows an example of a bar chart. Note that \our's attention is driven toward the dark-green bars due to the question asking about the \emph{average} for a certain category (\emph{secondary education}). 
As observed, CRCT attends intuitive features and spatial locations according to the questions asked. For more examples see the suppl. material.

\section{Summary and Discussion}
\label{sec:summary}
In this paper we argue that the simplicity of Chart Question Answering (CQA) associated with lack of realistic chart content and question types, has lead previous methods to omit the regression task. The recent PlotQA work \cite{plotqa} addresses these shortcomings, suggesting a remedy via a new large scale and diverse dataset, as well as a new model. We hereby suggest a bimodal framework for CQA that leverages the natural lingual inter-relations between different chart elements and introduce a novel unified classification-regression head. Our explainability visualizations shed light on question-chart understanding of our model.  

We evaluate our method on the PlotQA and FigureQA datasets, significantly outperforming the PlotQA model.
We further compare our method to a previous classification based method of PReFIL, that reached SoTA results on FigureQA (also high performing on DVQA) observing a strong drop in performance when tested on more challenging datasets such as PlotQA-D. We argue that the edge of our method is not in classification but rather on the combined classification regression tasks with natural lingual relations that exist in real CQA case.

However, some limitations still remains, such as sensitivity to 
color combinations and non-linear axis scales. Although we reach a comparable result to PReFIL on FigureQA, we noticed deterioration in results when the test and train colors are different. We relate this limitation to the detector representation learning, including the color attributes from the charts and relying on them to distinguish between the plots in a chart. In practice, this limitation can be overcome by extending the (synthetic) dataset to contain more colors.

In future work we intend to relax the need for full chart annotations, and tackle the efficiency of the training.
With PlotQA opening the door again toward chasing human performance in chart comprehension, we hope this paper will encourage researchers to take this challenge.

\paragraph{\bf Acknowledgments:} 
We thank Or Kedar and Nir Zabari for their assistance in parts of this research. We thank PlotQA \cite{plotqa} authors for sharing additional breakdowns.
This work was supported in part by the Israel Science Foundation (grant 2492/20).
%
%
\bibliographystyle{splncs04}
\bibliography{egbib}

\newpage
\appendix
\section*{\LARGE{Supplementary Material}}
In the supplementary material we elaborate on some results shown in the main paper, as well as present new ones. We start with further elaboration on the characteristics of Chart Question Answering (CQA) benchmarks in \cref{sec:public_cqa}, showing the shortcomings of previous public datasets such as FigureQA\cite{figureqa} and DVQA\cite{dvqa} by examples. Next, we elaborate on PlotQA\cite{plotqa} question type distribution, presenting the richness, realistic lingual relations, as well as regression requirement in this dataset. In \cref{sec:acc_metric} we show further justifications for the new accuracy metric. We show additional explainability results in \cref{sec:additional_explainability} to strengthen the reasoning process in our results. Next, we show our results on DVQA, emphasizing the shortcomings of this dataset to showcase our method.
In \cref{sec:lang_robust}, we show a new experiment for language robustness and compare our CRCT model with the PReFIL\cite{prefil} which showed high performance in previous benchmarks. Finally, In \cref{sec:new_gen_sample}, we run our model on a newly generated example (not from PlotQA) as a single demo case.

\section{Model Architecture}
In this section we provide some equations in order to further clarify our model descriptions in the paper. Let us denote the Query, Key and Value for each branch at certain block as $Q_v, K_v, V_v \in \mathbb{R}^{n_v\times d}$ and $Q_t, K_t, V_t \in \mathbb{R}^{n_t\times d}$ corresponding to the visual and textual branch respectively. Each branch is attended by the other, as the following:
\begin{eqnarray}
    z_t = attn_d(Q_v, K_t, V_t) := \mathit{softmax}(\frac{Q_v {K_t}^T}{\sqrt{d}}) V_t \in \mathbb{R}^{n_t\times d} \\
    z_v = attn_d(Q_t, K_v, V_v) := \mathit{softmax}(\frac{Q_t {K_v}^T}{\sqrt{d}}) V_v \in \mathbb{R}^{n_v\times d}
\end{eqnarray}
    The co-encoder output is followed by a regular self-attention encoder, namely:
\begin{eqnarray}
    \forall i\in \{t, v\}~~O_{i} = attn_d(Q(z_i), K(z_i), V(z_i))
\end{eqnarray}
Note that $Q_t, Q_v$ are exchanged, to allow interaction between different modalities. Then the outputs of each branch, $O_{t}, O_{v}$, are fed to the next co-transformers block in their proper branch. Finally, the resulting $h_{v0} , h_{w0}$ pooling tokens from the last layer are used for predicting if the concatenated answer is aligned (C) and the answer numeric value, R: 
\begin{eqnarray}
Loss_{CLS} = BCELoss(M_{CLS}(h_{w0} * h_{v0}), C)\\
Loss_{REG} = L1(M_{REG}([h_{w0};h_{v0}]), R)
\end{eqnarray}
We train our model with the combined loss:
\begin{equation}
    Loss = \lambda_1 \cdot Loss_{CLS} + \lambda_2 \cdot Loss_{REG}
\end{equation}
We find $\lambda_1 = \lambda_2 = 1$ to be effective.
\section{Characteristics of public CQA datasets}
\label{sec:public_cqa}
\begin{figure*}[!ht]
\minipage{0.25\textwidth}
  \subfloat[FigureQA]{\includegraphics[width=\linewidth]{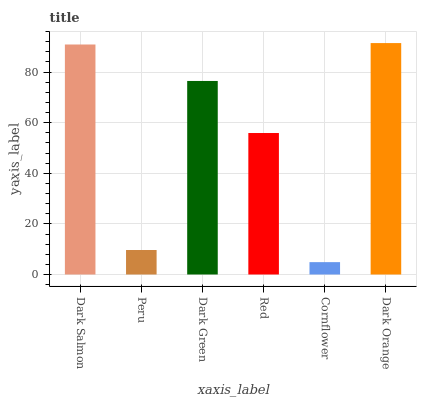}\label{fig:figureqa_sample_1}}
  \\ \subfloat[FigureQA]{\includegraphics[width=\linewidth]{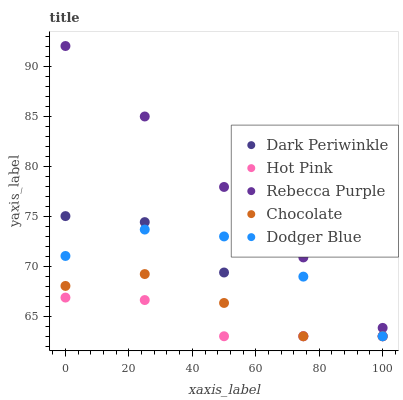}\label{fig:figureqa_sample_2}}
 
\endminipage\hfill
\minipage{0.25\textwidth}
  \subfloat[DVQA]{\includegraphics[width=\linewidth]{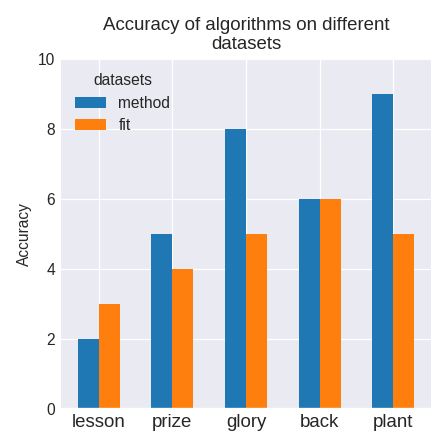}\label{fig:dvqa_sample_1}}
  \\ \subfloat[DVQA]{\includegraphics[width=\linewidth]{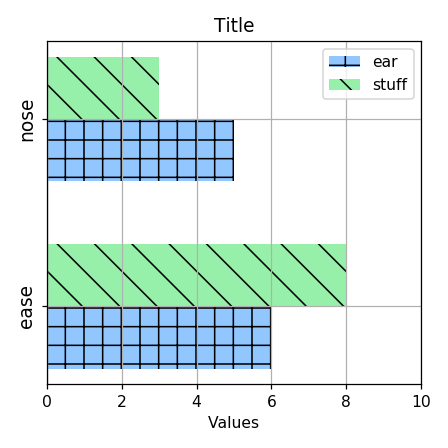}\label{fig:dvqa_sample_2}}
  
\endminipage\hfill
\minipage{0.5\textwidth}%
\subfloat[PlotQA]{\includegraphics[width=\linewidth]{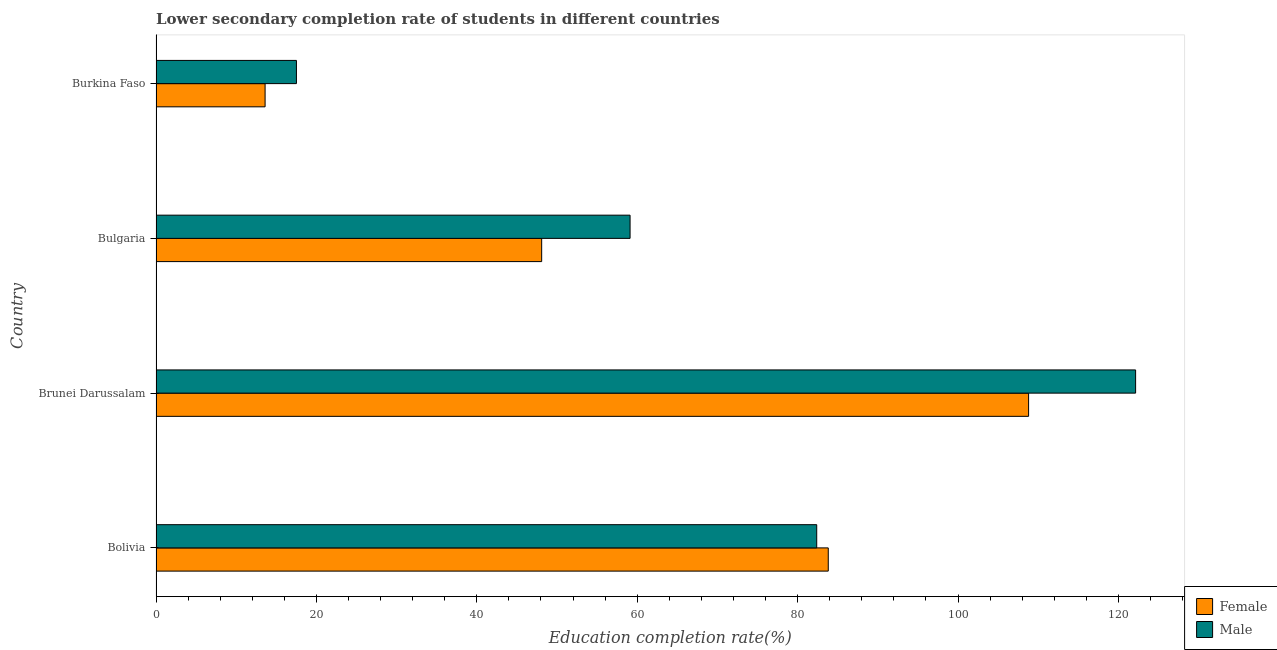}\label{fig:plotqa_sample_1}}
 \\ \subfloat[PlotQA]{\includegraphics[width=\linewidth]{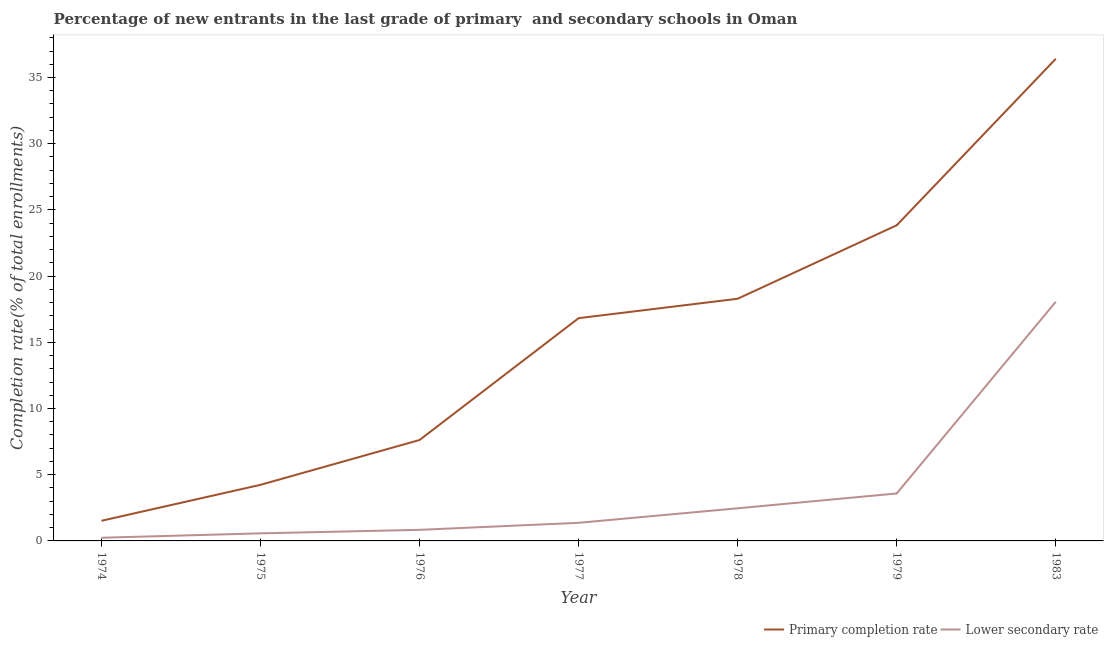}\label{fig:plotqa_sample_2}}
    
\endminipage
\caption{Examples of charts from the FigureQA, DVQA, and PlotQA datasets. FigureQA charts - (a) and (b) lack any diversity in title and axis labels as well as the plot labels. In DVQA - (c) and (d) Random phrases and words are used in the chart text, resulting in lack of natural semantic relations between the different textual elements. These drawbacks are addressed in PlotQA, where the charts are taken from real world data, as shown in (e) and (f). Zoom in for better visibility.}
\label{fig:datasets_figs}
\end{figure*}

In this section we elaborate on the characteristics of different chart datasets previously used for Chart Question Answering (CQA) methods and further justify the choice of PlotQA as our main benchmark dataset. To this end, we present examples of charts from FigureQA, DVQA, and PlotQA in \cref{fig:datasets_figs}. This figure demonstrates the fixed templates and degenerate lingual forms used in two previous datasets. In FigureQA, the title, x-axis label and y-axis label are fixed in all the charts. Additional template pattern in FigureQA includes, the legend markers of the plots (\eg, bars or scatter plots) named after their color. This pattern of naming is redundant throughout the entire dataset and is strictly used in the associated questions (see \cref{fig:figureqa_sample_1} \& \ref{fig:figureqa_sample_2}). DVQA alleviates part of these shortcomings, yet with random words used as legend or bar labels, as shown in \cref{fig:dvqa_sample_1} \& \ref{fig:dvqa_sample_2}. DVQA is limited to a single chart type and further introduces degenerated lingual forms that are unlikely to appear in a realistic chart. Note for instance, the word ``Title'' appearing as the title of the chart in \cref{fig:dvqa_sample_2}.

In \cref{fig:plotqa_sample_1} \& \ref{fig:plotqa_sample_2} we show examples from the PlotQA dataset.
To the best of our knowledge this is the most realistic dataset publicly available to date. In addition to it's size and diversity (see Table 1, main paper) it is the only dataset that satisfies all the following terms: 1) Publicly available; 2) Fully annotated to train a detector; 3) Includes multiple chart types; 4) Charts with natural language patterns and relations; 5) Questions that demand regression.

These dataset characteristics are strongly related to the performance drop that was recently reported on PlotQA dataset in \cite{plotqa}, and discussed in the paper. We further discuss additional factors, such as lack of regression required questions in previous benchmarks, in the main paper. 
In \cref{table:dataset_accuracies} we summarize the performance of several recently published methods against the existing datasets.
\begin{table}[ht]
\caption{Accuracy of different methods on existing datasets. Note the significant drop in accuracy on PlotQA dataset (PlotQA-D).  $^*$ our evaluation of PReFIL method on PlotQA-D}
\begin{center}
\begin{tabular}{@{}lcccc@{}}
\toprule
\multicolumn{1}{c}{\begin{tabular}[c]{@{}c@{}}Method / Dataset\end{tabular}} & FigurQA-D & DVQA-D & LeafQA-D & PlotQA-D      \\ \midrule
PReFIL\cite{prefil}                                                            & 93.26     & 96.4   & -        & \textbf{10.36$^*$} \\
STL-CQA\cite{chartqa}                                                          & -         & 97.43  & 92.22    & -              \\
LEAF-QA\cite{chaudhry2019leafqa}                                               & 81.15     & 72.8   & 67.42    & -              \\
PlotQA-M\cite{plotqa}                                                          & -         & 58.78  & -        & \textbf{22.52} \\ \bottomrule
\end{tabular}
\end{center}
\label{table:dataset_accuracies}
\end{table}

\section{PlotQA Data Distribution}
\begin{figure*}[ht]
  \centering
  \subfloat[Structural]{\includegraphics[width=0.5\textwidth]{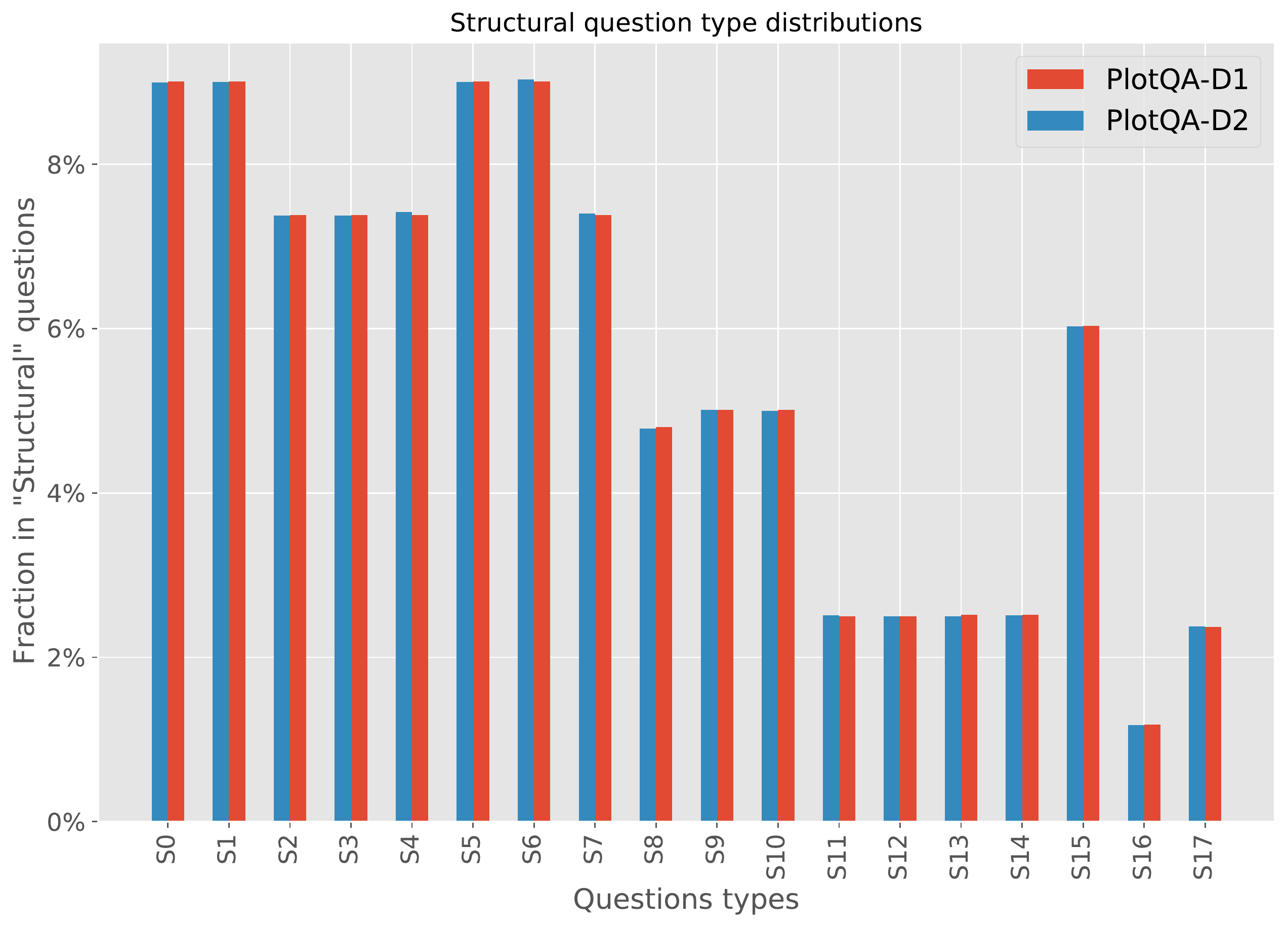}\label{fig:data_dist_structural}}
  \hfill
  \subfloat[Data Retrieval]{\includegraphics[width=0.5\textwidth]{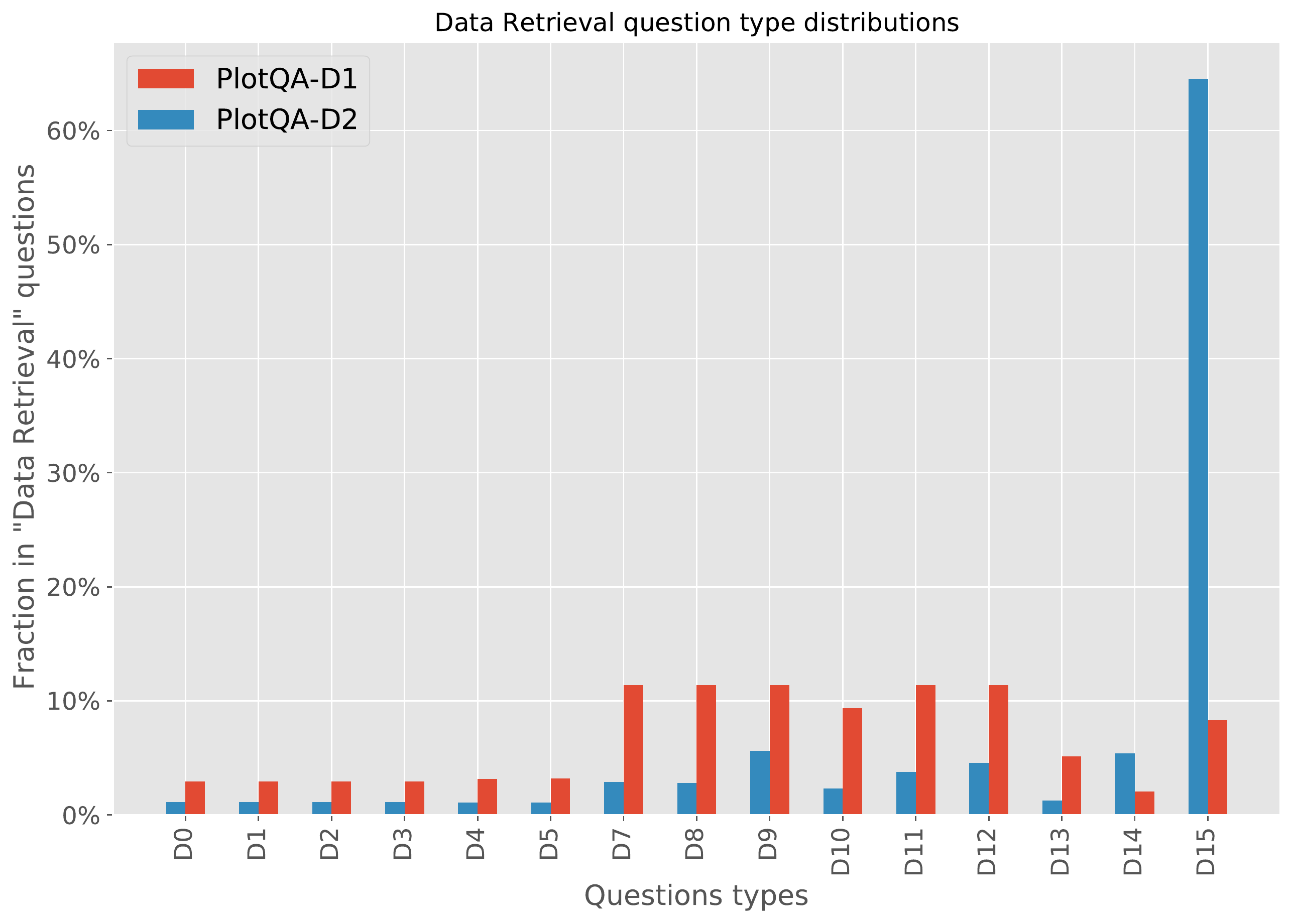}\label{fig:data_dist_data}}
   \\
  \subfloat[Reasoning]{\includegraphics[width=0.6\textwidth]{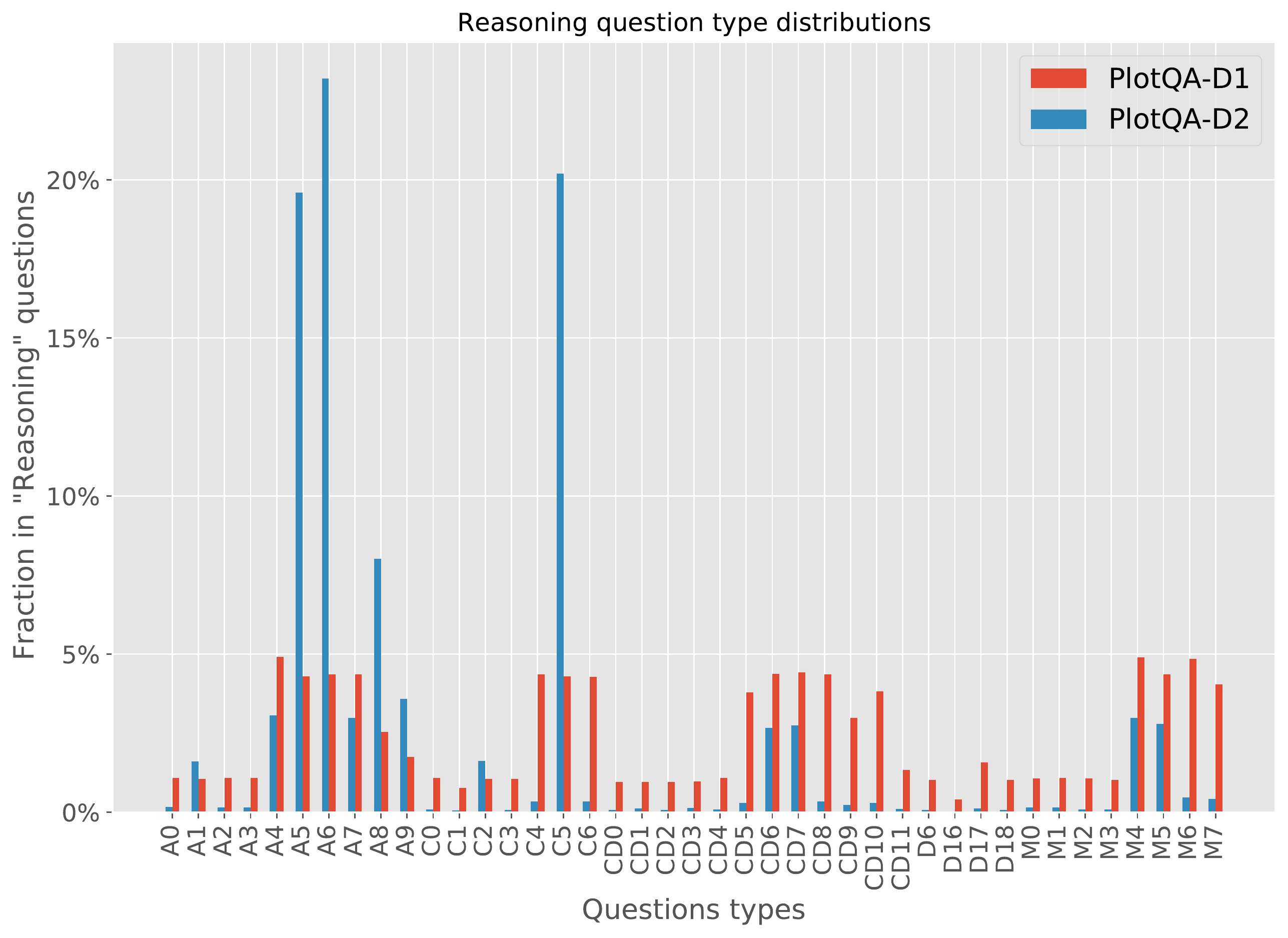}\label{fig:data_dist_reasoning}}
    \caption{PlotQA-D1 and D2 question type distributions. While in Structural questions the distribution is similar, in Data Retrieval and Reasoning questions, PlotQA-D2 is skewed towards few specific templates, which require a regression answer. }
\label{fig:plotqa-v1_v2_distribution}
\end{figure*}

The PlotQA dataset suggests two benchmarks, which we refer to as PlotQA-D1 and PlotQA-D2 (see the paper for chart and Q\&A breakdown). Both datasets share the same chart images. However, PlotQA-D1 is a subset of PlotQA-D2, with the latter having $\times$3.5 more Q\&As. \cref{table:plotqa_question_distribution} shows the question type distributions for each benchmark with \cref{fig:plotqa-v1_v2_distribution} depicting distributions of question templates in each question category, Structural (S), Data Retreival (D) and Reasoning (R). PlotQA-D1 introduces a relatively uniform distribution over the question templates, while PlotQA-D2 distribution is strongly skewed by a large number of questions requiring regression (with non-integer answers). PlotQA-D2 was designed to showcase the capability of a method on handling regression, a highly practical task and a strong shortcoming of previous datasets. The results reported in the paper demonstrate that \our outperforms previous methods on \emph{both} of these benchmarks.


\begin{table}[ht]
\begin{center}
\caption{Distribution over different question categories in PlotQA benchmarks}
\vspace{5mm}
\begin{tabular}{@{}lccc@{}}
\toprule
Data Ver. & \multicolumn{1}{l}{Structural} & \multicolumn{1}{l}{Data Retrieval} & \multicolumn{1}{l}{Reasoning} \\ \midrule
PlotQA-D1 & 30.41\%                        & 24.01\%                            & 45.58\%                       \\
PlotQA-D2 & 4.3\%                          & 13.74\%                            & 81.96\%                       \\ \bottomrule
\end{tabular}
\label{table:plotqa_question_distribution}

\end{center}

\end{table}


\section{Accuracy Metric}
\label{sec:acc_metric}
\begin{figure*}[ht]
  \centering
  \subfloat[A visual comparison between values and their error ranges, according to $\pm5\%$ ratio metric (red) and $\pm\nicefrac{1}{2}$ sub-tick metric (green). The error tolerance of the error ratio measure depends on the ground truth value.]{\includegraphics[width=0.4\textwidth]{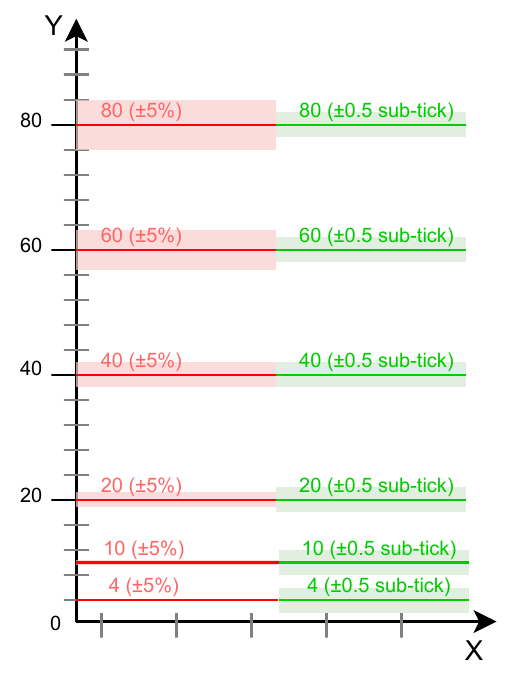}\label{fig:tick_vs_ratio_metric}}
  \hfill
  \subfloat[Distribution of regression errors by two different metrics, the $\pm5\%$ tolerance, in blue, and our suggested tick metric (Sec. 5 in the paper), in red. Note the peak in errors near zero, while the fixed tick based tolerance results nearly uniform distribution.]{\includegraphics[width=0.5\textwidth]{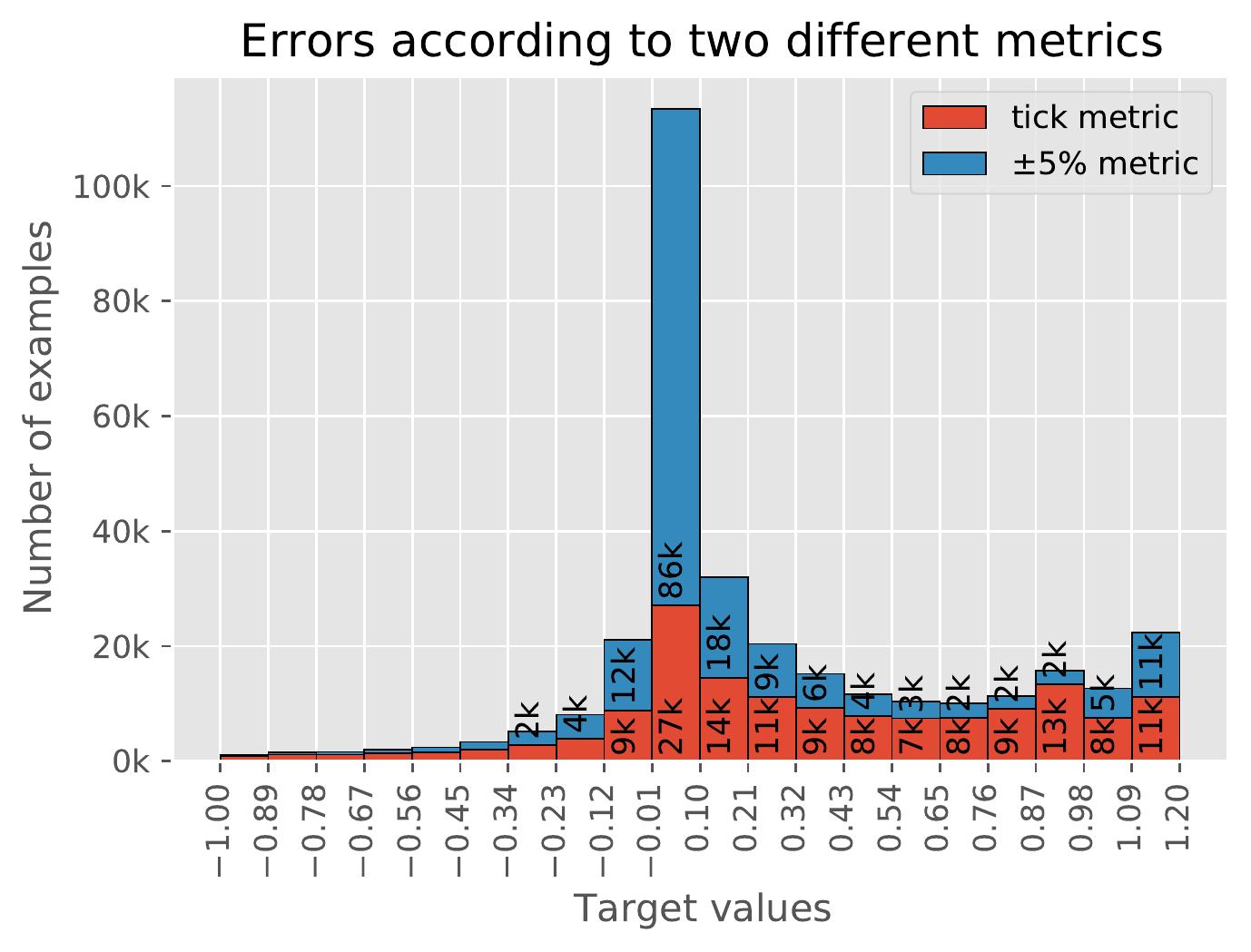}\label{fig:regressor_metric_comparison}}

  \caption{Ratio vs Sub-tick metric.}
\label{fig:ratio_vs_subtick}
\end{figure*}

In Fig. \ref{fig:tick_vs_ratio_metric} we graphically visualize the dependency of the error tolerance on the ground truth value for the error ratio measure, in contrast to a fixed tolerance in the tick-based error, as suggested in our paper. Note the vanishing of the tolerance as the true value goes to zero (and vice versa).

This bias in the ratio based measure drives the errors to accumulate near zero as we show in Fig \ref{fig:regressor_metric_comparison}. This figure presents a comparison between the error ratio measure and the suggested tick based error, for CRCT on PlotQA-D1, showing the bias in ratio based tolerance. We observe a relatively uniform error distribution on the ticked based alternative, as desired.  

\section{Additional Explainability Examples}
\label{sec:additional_explainability}
\begin{figure*}[htpb]
\begin{center}
   \includegraphics[width=0.7\textwidth]{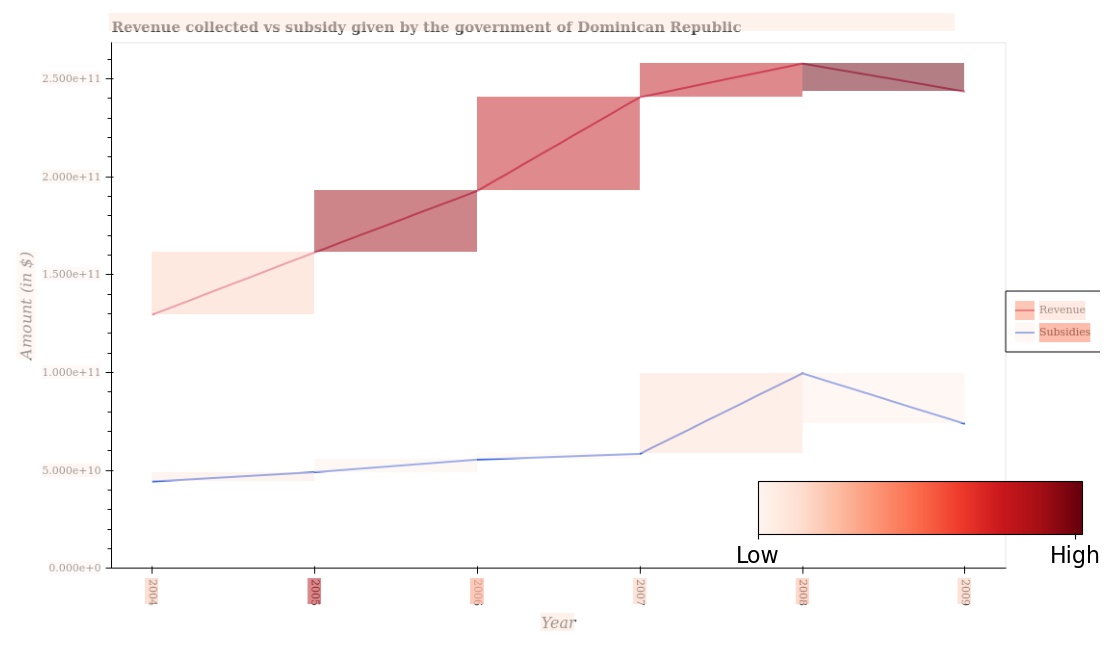}
\end{center}
   \caption{Explainability visualizations for a PlotQA test sample. Q: {\it Is the amount of revenue collected in 2005 less than that in 2008?} ground truth: Yes. \our: \textcolor{ForestGreen}{Yes}. Note the high attention on the correct plot between the two. The font sizes are from the dataset source.}
\label{fig:visualization_639418_}
\end{figure*}

In this section we show more visualization examples on CRCT explainability using {\it Captum} visualization tool (see Sec.~7 in the paper). All examples are drawn from the test set. In \cref{fig:visualization_639418_} we present a case with two line-plots in a chart. Note how the model attends to the correct plot among the two (hot bounding boxes) when asked about the {\it revenue}.

\begin{figure*}[htpb]
\begin{center}
   \includegraphics[width=0.9\textwidth]{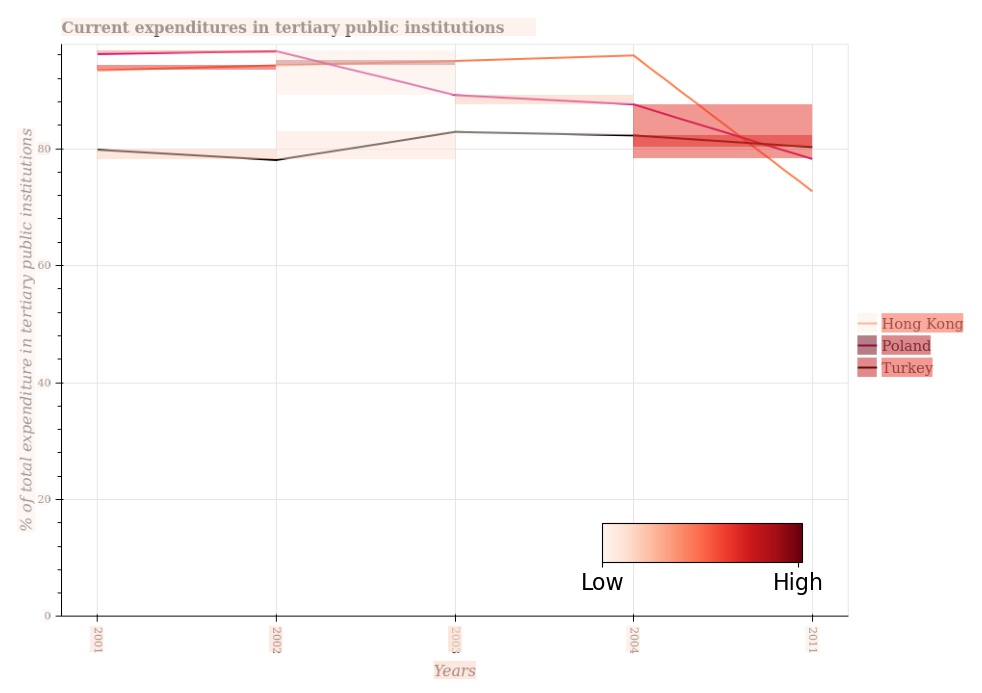}
\end{center}
   \caption{Explainability visualizations for a PlotQA test sample. Q: {\it How many lines intersect with each other?}, Ground truth: 3. \our: \textcolor{ForestGreen}{3}. Note the hot spots at the intersection points.}
\label{fig:visualization_690053}
\end{figure*}

\cref{fig:visualization_690053} shows an example of semantic understanding. Asked about the {\it intersection}, CRCT mostly attends to the two intersection points in the chart.

\begin{figure*}[htpb]
\begin{center}
   \includegraphics[width=0.8\textwidth]{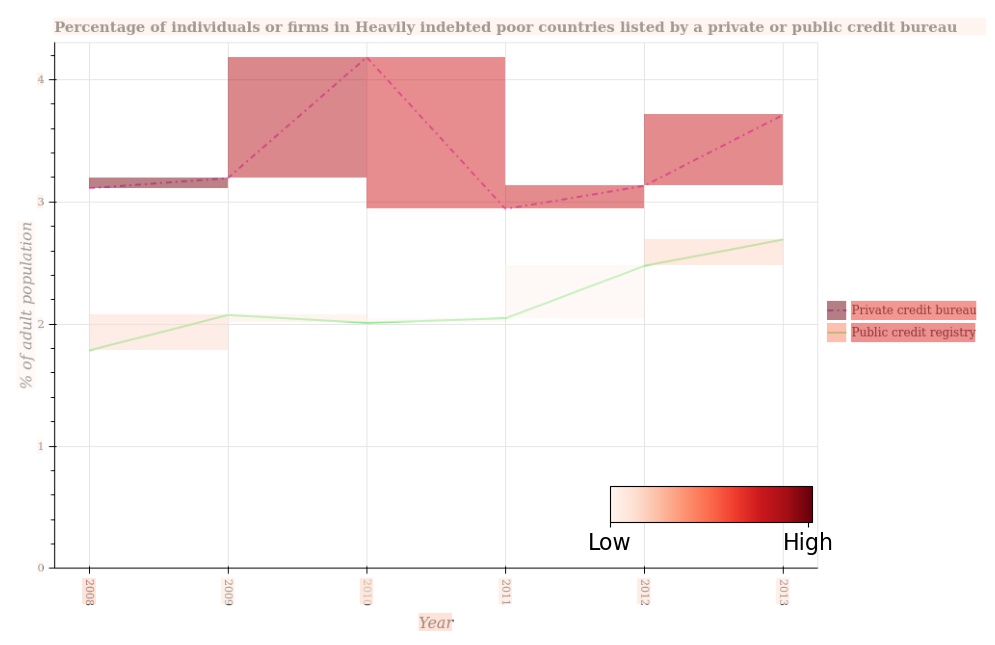}
\end{center}
   \caption{Explainability visualizations for a PlotQA test sample. Q: {\it What is the average percentage of firms listed by \textbf{private} credit bureau per year?} Ground truth: 3.379. \our: \textcolor{ForestGreen}{3.295} (Error: \textcolor{Red}{-2.49\%}). Note how the model attends the correct plot among the two, with ''hot" bounding boxes over all the plot due to {\it average} request.}
\label{fig:visualization_668158}
\end{figure*}

\cref{fig:visualization_668158} shows another multi-plot chart. Here, the model correctly finds the {\it private credit} line plot as more influential to the question asked. Furthermore, asking about the {\it average} value drives CRCT to attend to all the plot elements corresponding to the {\it private credit} label.

\begin{figure*}[ht]
\begin{center}
   \includegraphics[width=0.8\textwidth]{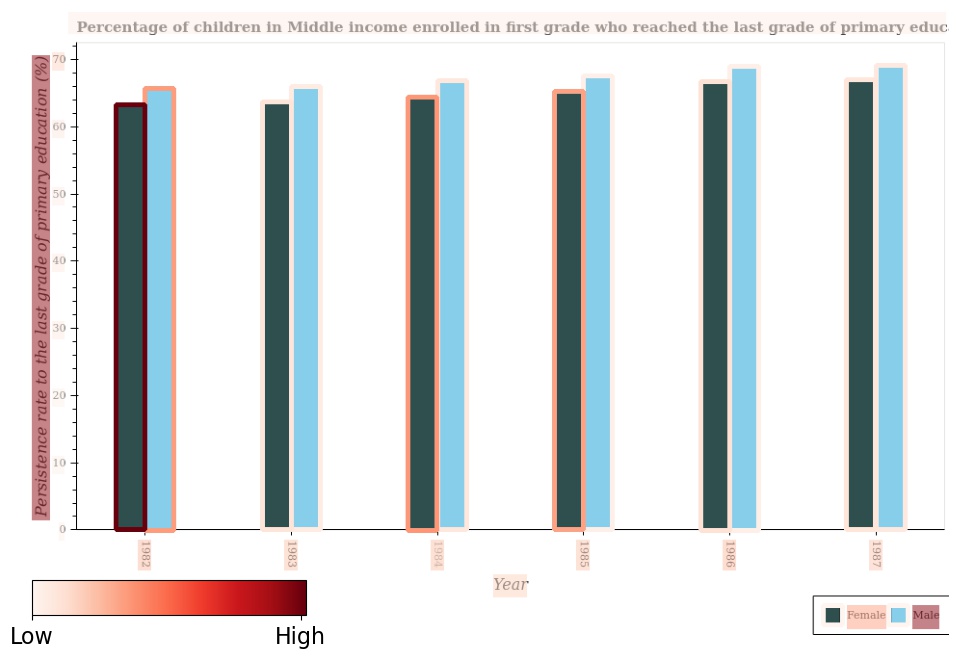}
\end{center}
   \caption{Explainability visualizations for a PlotQA test sample. Q: {\it In which year was the persistence rate of female students minimum?}, Ground truth: 1982. \our: \textcolor{ForestGreen}{1982}. For better visibility, we overlay the visualization as colored bounding box around the bars. Note how green bars related to {\it Female} achieve higher attention with the correct bar receiving the highest attention.}
\label{fig:visualization_864287}
\end{figure*}
In the next example in \cref{fig:visualization_864287} we show a bar chart. Although the bars are very close in their heights (values), the relevant bar, with the minimum value gets the highest attention, leading to the correct answer.


\section{Result on DVQA}
DVQA dataset is limited by 1) Single chart type (bar charts), 2) Lack of natural lingual text in the chart (see \cref{sec:public_cqa}), 3) Answers appearing as a-priori known classes, eliminating the need for regression. This dataset further lacks the important legend marker annotation needed to train our detector. The importance of this object is clearly shown in the explainability examples in \cref{sec:additional_explainability} (and in the paper), where legend markers are frequently highlighted, allowing \our to correspond to the correct plot/bar in the chart. The results of our CRCT model are shown in Table \ref{table:dvqa_results}.
Despite the limitations above, and errors involved in our heuristic annotation, we achieve a reasonable performance of 82.14\%, ranked 3rd, on this benchmark and far beyond PlotQA-M that achieves 57.99\% . Note that to showcase the strong limitation and existing performance saturation on DVQA we evaluate PReFIL, that reaches almost perfect performance on DVQA (96.37\%), on the new PlotQA dataset (see results in the paper).
\begin{table}[ht]
\centering
\caption{Results on DVQA dataset. ${CRCT}_p$ indicates the \our model with a detector that was trained with {\it partial} bounding box annotations.}
\vspace{5mm}
\begin{tabular}{@{}l|llllll@{}}
\toprule
Method & SANDY\cite{dvqa} & PlotQA-M\cite{plotqa} & LEAF-Net\cite{chaudhry2019leafqa} & ${CRCT}_p$~& PReFIL\cite{prefil} & STL-CQA\cite{chartqa} \\ \midrule
Accuracy & \multicolumn{1}{c}{56.48} & \multicolumn{1}{c}{57.99} & \multicolumn{1}{c}{72.72} & \multicolumn{1}{c}{82.14} & \multicolumn{1}{c}{96.37} & \multicolumn{1}{c}{{\bf 97.35}} \\ \bottomrule
\end{tabular}
\label{table:dvqa_results}
\end{table}

\section{Language Robustness}
\label{sec:lang_robust}
In this section we demonstrate the robustness under natural lingual variations of the CRCT model. Our transformer based model allows initialization with pretrained BERT \cite{devlin2019bert} on a large language corpus such as Wikipedia. \our is further trained downstream on all the textual elements in the chart, without any heuristics, such as string replacements (see the paper). This is in contrast to previous methods, often using LSTM, based only on the chart dataset vocabulary. We further compare the CRCT robustness with PReFIL \cite{prefil}, where a LSTM is used for question encoding. In Tables \ref{table:x_axis_variations}, \ref{table:how_many_bars_variations} and \ref{table:d15_variations} we present question rephrasing on test figures. Each variation is a new manual phrasing of the original template. Note that the original template is the only one appearing in train set. Tables \ref{table:x_axis_variations}-\ref{table:d15_variations} show that while on the template question PReFIL gives the correct answer (indicated in green), it is mostly wrong (indicated in red) after question rephrasing. CRCT however is more robust to phrasing for various question types \eg data retrieval, and regression.
\begin{table*}[htpb]
\caption{Question rephrasing, for \cref{fig:visualization_690053}. Each variation was manually rephrased and never seen in train, except the original version. Note the high sensitivity of PReFIL to different question phrasings.}
\center
\resizebox{1\textwidth}{!}{%

\begin{tabular}{clcc}
\hline
\multicolumn{1}{l}{Var.} & Question & CRCT & PReFIL \\ \hline
Original & What is the label or title of the X-axis ? & \textcolor{ForestGreen}{Years} & \textcolor{ForestGreen}{Years} \\ \hline
\#1 & What's the name of the X-axis? & \textcolor{ForestGreen}{Years} & \textcolor{Red}{40} \\
\#2 & What is the label or title of the horizontal axis? & \textcolor{ForestGreen}{Years} & \textcolor{Red}{0} \\
\#3 & What is the x label of the plot? & \textcolor{ForestGreen}{Years} & \textcolor{Red}{2011} \\
\#4 & The x-label of the figure? & \textcolor{Red}{2004} & \textcolor{Red}{0} \\
\#5 & What's the figure's x-axis label? & \textcolor{ForestGreen}{Years} & \textcolor{Red}{2011} \\
\#6 & Give me the x-axis label & \textcolor{ForestGreen}{Years} & \textcolor{Red}{2011} \\
\#7 & What the x-axis represents? & \textcolor{Red}{No} & \begin{tabular}[c]{@{}c@{}}\textcolor{Red}{\% of total expenditure}\\ \textcolor{Red}{in tertiary public institutions}\end{tabular} \\
\#8 & What is the label of X? & \textcolor{ForestGreen}{Years} & \textcolor{Red}{2011} \\
\#9 & X-label? & \textcolor{Red}{2011} & \textcolor{Red}{2011} \\ \hline
\end{tabular}
}

\label{table:x_axis_variations}

\end{table*}

\begin{table*}[htpb]
\caption{Question rephrasing, for \cref{fig:plotqa_sample_1}. Each variation was manually rephrased and never seen in train, except the original version. Note the high sensitivity of PReFIL to different question phrasings.}
\center
\resizebox{0.8\textwidth}{!}{
\begin{tabular}{@{}clcc@{}}
\toprule
\multicolumn{1}{l}{Var.} & Question & CRCT & PReFIL \\ \midrule
Original & How many different coloured bars are there ? & \textcolor{ForestGreen}{2} & \textcolor{ForestGreen}{2} \\ \midrule
\#1 & How many bar colors are there? & \textcolor{ForestGreen}{2} & \textcolor{ForestGreen}{2} \\
\#2 & How many colors of bars can you see? & \textcolor{ForestGreen}{2} & \textcolor{Red}{1} \\
\#3 & Coloured bars? & \textcolor{Red}{No} & \textcolor{Red}{Bolivia} \\
\#4 & How many colors paints each group of bars? & \textcolor{ForestGreen}{2} & \textcolor{Red}{Bolivia} \\
\#5 & How many colors are there? & \textcolor{ForestGreen}{2} & \textcolor{ForestGreen}{2} \\
\#6 & How many different bars exists in each group? & \textcolor{ForestGreen}{2} & \textcolor{Red}{0} \\
\#7 & Colors in each group? & \textcolor{Red}{No} & \textcolor{Red}{0} \\
\#8 & How many colors? & \textcolor{ForestGreen}{2} & \textcolor{Red}{0} \\
\#9 & Give me the size of each group of bars & \textcolor{Red}{Bolivia} & \textcolor{ForestGreen}{2} \\ \bottomrule
\end{tabular}
}

\label{table:how_many_bars_variations}
\end{table*}

\begin{table*}[htpb]
\center
\caption{Question rephrasing, for \cref{fig:plotqa_sample_2}. Each variation was manually rephrased and never seen in train, except the original version. Note that this question requires regression. In every variation therefore the $\langle R\rangle$ token was chosen in \our's hybrid prediction head, leading to the regression value shown as an answer. The values in green and red are correct and wrong answers respectively. Values in blue present the deviation from the true value. Note the high sensitivity of PReFIL to different question phrasings}
\resizebox{1\textwidth}{!}{

\begin{tabular}{clll}
\hline
\multicolumn{1}{l}{Var.} & Question & CRCT & PReFIL \\ \hline
Original & Across all years, what is the maximum completion rate in primary schools ? & \textcolor{ForestGreen}{36.618} \textcolor{Blue}{($+0.57$\%)} & \textcolor{ForestGreen}{35} \textcolor{Blue}{($-3.87$\%)} \\ \hline
\#1 & What's the maximum primary completion? & \textcolor{ForestGreen}{36.625} \textcolor{Blue}{($+0.59$\%)} & \textcolor{Red}{0} \textcolor{Blue}{($-100$\%)} \\
\#2 & What is the maximal rate of primary school completion, over the years? & \textcolor{ForestGreen}{36.305} \textcolor{Blue}{($-0.288$\%)} & \textcolor{Red}{No} \\
\#3 & Across all years, what is the maximum primary school completion rate? & \textcolor{ForestGreen}{36.613} \textcolor{Blue}{($+0.56$\%)} & \textcolor{Red}{0} \textcolor{Blue}{($-100$\%)} \\
\#4 & Over the years, what is the highest primary school completion rate? & \textcolor{ForestGreen}{36.502} \textcolor{Blue}{($+0.25$\%)} & \textcolor{Red}{No} \\
\#5 & What is the maximum completion rate in primary schools across all years? & \textcolor{ForestGreen}{36.635} \textcolor{Blue}{($+0.618$\%)} & \textcolor{Red}{0} \textcolor{Blue}{($-100$\%)} \\
\#6 & In primary schools, what is the highest completion rate across all years? & \textcolor{ForestGreen}{36.53} \textcolor{Blue}{($+0.33$\%)} & \textcolor{Red}{3} \textcolor{Blue}{($-91.76$\%)} \\
\#7 & The maximum completion rate in primary schools is what - across all years? & \textcolor{ForestGreen}{36.622} \textcolor{Blue}{($+0.58$\%)} & \textcolor{Red}{0} \textcolor{Blue}{($-100$\%)} \\
\#8 & Give me the maximum rate of primary completion over the graph & \textcolor{Red}{33.581} \textcolor{Blue}{($-7.77$\%)} & \textcolor{Red}{5} \textcolor{Blue}{($-86.27$\%)} \\
\#9 & Average the primary completion rate & \textcolor{Red}{15.065} \textcolor{Blue}{($-58.624$\%)} & \textcolor{Red}{No} \\ \hline
\end{tabular}
}

\label{table:d15_variations}
\end{table*}

\section{New Generated Example}
\label{sec:new_gen_sample}
\begin{figure*}[ht]
\begin{center}
   \includegraphics[width=0.8\textwidth]{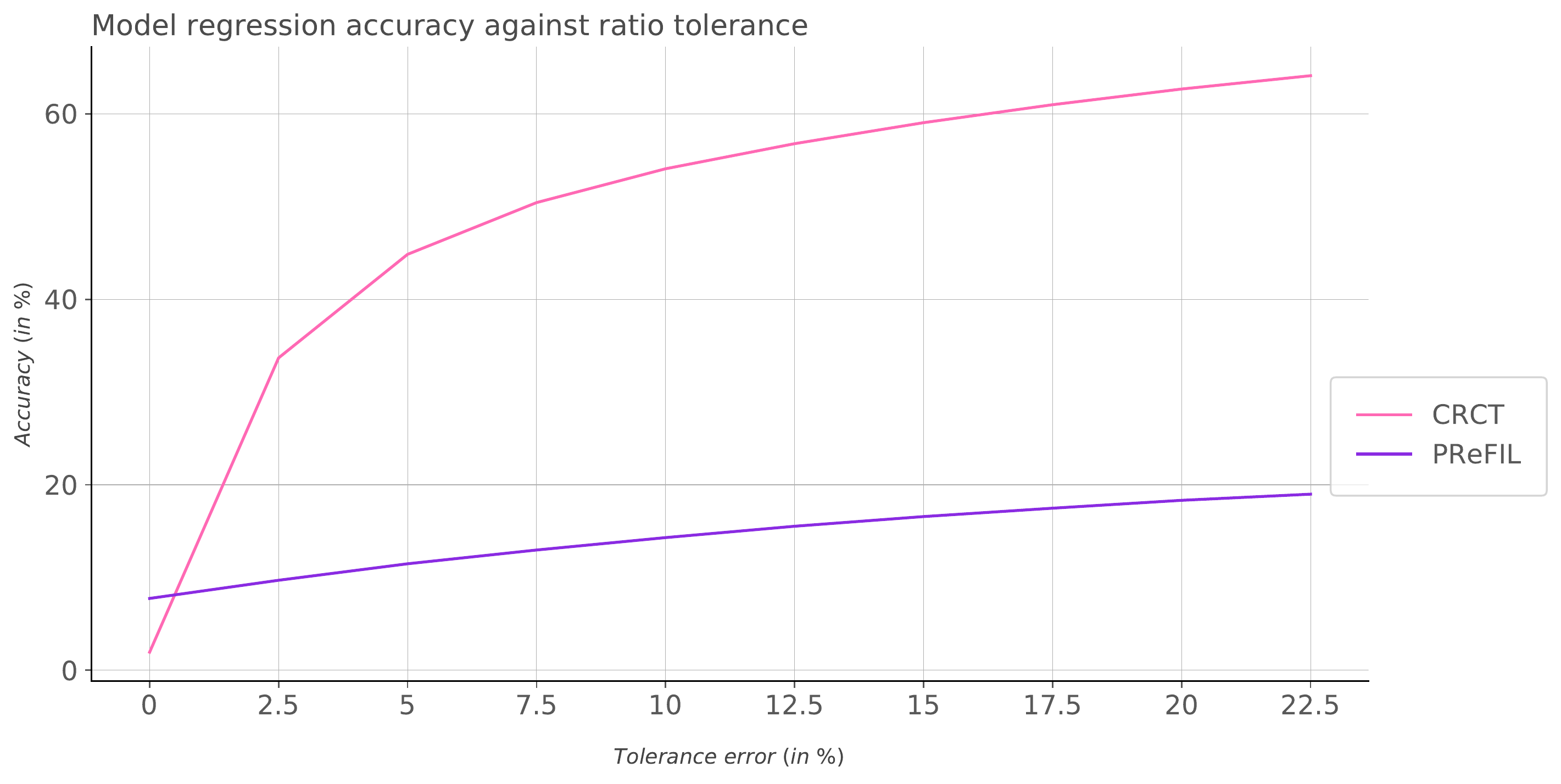}
\end{center}
   \caption{We insert into CRCT a result from our paper showing the regression accuracy of our model against PReFIL. We pose the following question: {\it In 2.5\% tolerance error, what is the difference between the accuracy of \textbf{CRCT} and \textbf{PReFIL}?}. Ground truth: 23.978. \our: \textcolor{ForestGreen}{23.865} (\textcolor{Red}{-0.47\%}).}
\label{fig:crct_prefil_comp_ratio}
\end{figure*}

We conclude by showing a result from an experiment in our study in Fig. \ref{fig:crct_prefil_comp_ratio}. To this end, we create a new chart showing our accuracy result compared to PReFIL. We now pose the following question to the model: {\it In 2.5\% tolerance error, what is the difference between the accuracy of CRCT and PReFIL?}. Although this figure was not part of the PlotQA dataset, we obtained an answer that deviates the true result only by 0.47\%. The robustness of CRCT is further illustrated here on handling unknown initials of the corresponding methods.

\clearpage
\newpage

\end{document}